\documentclass[3p]{elsarticle}

\usepackage[T1]{fontenc}

\makeatletter
\def\ps@pprintTitle{%
\let\@oddhead\@empty
\let\@evenhead\@empty
\def\@oddfoot{}%
\let\@evenfoot\@oddfoot}
\makeatother

\usepackage{graphicx}

\usepackage{amsmath,amsfonts}
\usepackage{mathtools}
\usepackage{amsthm}
\newtheorem{definition}{Definition}
\newtheorem{lemma}{Lemma}

\theoremstyle{definition}

\usepackage{booktabs}

\usepackage{url}
\newcommand{\e}[1]{\emph{#1}}
\newcommand{\bftab}{\fontseries{b}\selectfont}

\newcommand{\R}{\mathbb{R}}

\newcommand{\nin}{\notin}
\newcommand{\ol}[1]{\overline{#1}}
\newcommand{\ul}[1]{\underline{#1}}

\newcommand{\lra}{\longrightarrow}
\newcommand{\lrma}{\longmapsto}

\newcommand{\ra}{\rightarrow}
\newcommand{\rma}{\mapsto}

\RequirePackage{color}\definecolor{RED}{rgb}{1,0,0}

\begin{document}

\begin{frontmatter}

\title{A unified weighting framework for evaluating nearest neighbour classification}

\author[label1,label2]{Oliver Urs Lenz\corref{cor1}} %% Author name
\ead{o.u.lenz@liacs.leidenuniv.nl}
\cortext[cor1]{Corresponding author at: Leiden Institute of Advanced Computer Science, Leiden University, 2333 CA Leiden, The Netherlands.}
\author[label2]{Henri Bollaert} %% Author name
\ead{henri.bollaert@ugent.be}
\author[label2]{Chris Cornelis} %% Author name
\ead{chris.cornelis@ugent.be}

\affiliation[label1]{organization={Leiden Institute of Advanced Computer Science, Leiden University},%Department and Organization
            %addressline={},
            city={Leiden},
            postcode={2333 CA},
            %state={},
            country={The Netherlands}}
\affiliation[label2]{organization={Department of Applied Mathematics, Computer Science and Statistics, Ghent University},%Department and Organization
            %addressline={},
            city={Ghent},
            postcode={9000},
            %state={},
            country={Belgium}}

%% Abstract
\begin{abstract}
We present the first comprehensive and large-scale evaluation of classical (NN), fuzzy (FNN) and fuzzy rough (FRNN) nearest neighbour classification. We standardise existing proposals for nearest neighbour weighting with kernel functions, applied to the distance values and/or ranks of the nearest neighbours of a test instance. In particular, we show that the theoretically optimal Samworth weights converge to a kernel. Kernel functions are closely related to fuzzy negation operators, and we propose a new kernel based on Yager negation. We also consider various distance and scaling measures, which we show can be related to each other. Through a systematic series of experiments on 85 real-life classification datasets, we find that NN, FNN and FRNN all perform best with Boscovich distance, and that NN and FRNN perform best with a combination of Samworth rank- and distance-weights and scaling by the mean absolute deviation around the median ($r_1$), the standard deviation ($r_2$) or the semi-interquartile range ($r_{\infty}^*$), while FNN performs best with only Samworth distance-weights and $r_1$- or $r_2$-scaling. However, NN achieves comparable performance with Yager-$\frac{1}{2}$ distance-weights, which are simpler to implement than a combination of Samworth distance- and rank-weights. Finally, FRNN generally outperforms NN, which in turn performs systematically better than FNN.
\end{abstract}

%% Keywords

\end{frontmatter}

\section{Introduction}

Nearest neighbours (NN) \cite{fix51discriminatory} is one of the oldest and most widely used classification algorithms. It identifies the nearest neighbours (the most similar records) of a given test record in the training set, and predicts class scores based on the prevalence of each decision class among these neighbours. Despite the development of other successful classifiers, it remains an attractive option in many settings due to its simplicity \cite{cunningham21knearest} and good performance for many problems, especially when decision boundaries are irregular \cite{james23statistical}. Even when other algorithms achieve higher classification performance, NN may be preferable due to its interpretability: every prediction can be explained to a practitioner by comparing a test record with a handful of its nearest neighbours. NN and its variants are applied in such diverse domains as healthcare \cite{arian20protein,kour22visionbased,shahrestani23developing}, agriculture \cite{cosenza21comparison,gomezgil24vibrationbased}, geology \cite{martinmartin23using,suleymanov23spatial}, industry \cite{aslinezhad20turbine,konieczny21use,shijer24evaluating} and natural language processing \cite{khandelwal21nearest,kaminska23fuzzyanalysis,kaminska23fuzzyirony}, and have also been adapted to big data settings \cite{maillo20fast,lenz20scalable,shokrzade21novel}.

While NN is a relatively simple algorithm, it still requires setting a few hyperparameters. Even in its most basic form, we need to choose a distance measure and a method to rescale the attributes of the dataset. In addition, it is generally advisable to choose the number of neighbours $k \geq 1$ on which predictions are based. Choosing a larger value for $k$ increases the bias of the model and its tolerance for noisy training records, at the cost of reduced flexibility. Finally, there have been many proposals in the literature to weigh the contribution of the $k$ nearest neighbours of a test record differently.

In principle, it is possible to resolve these choices for any given problem by picking the combination of hyperparameter values that performs best on cross-validated training data. However, in practice, it is often more convenient to focus these efforts on the value $k$, and set the other hyperparameters to values that are known to be good enough. To aid this approach, it would be useful to have an idea which choices generally perform better than others.

Building on the traditional form of (weighted) NN described above, some authors have proposed further-reaching modifications that incorporate fuzzy set theory: fuzzy nearest neighbours (FNN) \cite{keller85fuzzy} and fuzzy rough nearest neighbours (FRNN) \cite{jensen08new}. FNN operates in a similar way to NN, but uses fuzzified class membership degrees of training records, while FRNN models each decision class as a fuzzy set, and calculates the membership degrees of a test record in these fuzzy sets. Like NN, these algorithms require a choice of distance measure, scale, number of neighbours $k$ and weighting scheme.

While there has been no shortage of new ideas for NN variants, it remains largely unclear which proposals work better in practice. The goal of this paper is to address this. Concretely, our contributions are the following:

\begin{itemize}
 \item We present a comprehensive overview of the different weighting proposals for NN, FNN and FRNN in the literature.
 \item We establish a common weighting framework in terms of kernel functions that unifies these proposals.
 \item In particular, we prove that the theoretically optimal rank-weights identified by Samworth \cite{samworth12optimal} converge to a specific kernel function as $k$ increases.
 \item We show that kernel functions are closely related to fuzzy negation operators. In particular, the weights proposed by Gou et al. \cite{gou12new} correspond to Sugeno negation. Inspired by this, we propose our own weighting kernel based on Yager negation, which has a distinct contour from existing weighting proposals.
 \item We introduce the concept of the Minkowski $p$-radius $r_p$ of a dataset, and show that $r_2$ and $r_{\infty}$ are, respectively, the standard deviation and the half-range, two commonly used measures of dispersion. We propose that $r_1$, the mean absolute deviation around the median, should also be considered as a scaling measure, in particular because it is less sensitive to outliers.
 \item We conduct a large-scale experiment on 85 real-life classification datasets, comparing distance and scaling measures and weighting kernels. To the best of our knowledge, this is the first large-scale evaluation of all NN weighting proposals, and the first direct comparison of FRNN with NN and FNN.
\end{itemize}

In Section~\ref{sec_nn_background}, we present an overview of the literature on nearest neighbour classification variants. Next, we present our own proposals in Section~\ref{sec_proposals}. We then describe our experimental setup (Section~\ref{sec_nn_experimental_setup}) and present the results (Section~\ref{sec_nn_results}), before concluding (Section~\ref{sec_nn_conclusion}).

\section{Background}
\label{sec_nn_background}
In this section, we briefly review the existing literature on nearest neighbour weighting, as well as previous experimental comparisons.

\subsection{Weighted nearest neighbour classification}
\label{sec_weighted_nearest_neighbour_classification}

\begin{figure}
\centering
\includegraphics[width=\linewidth]{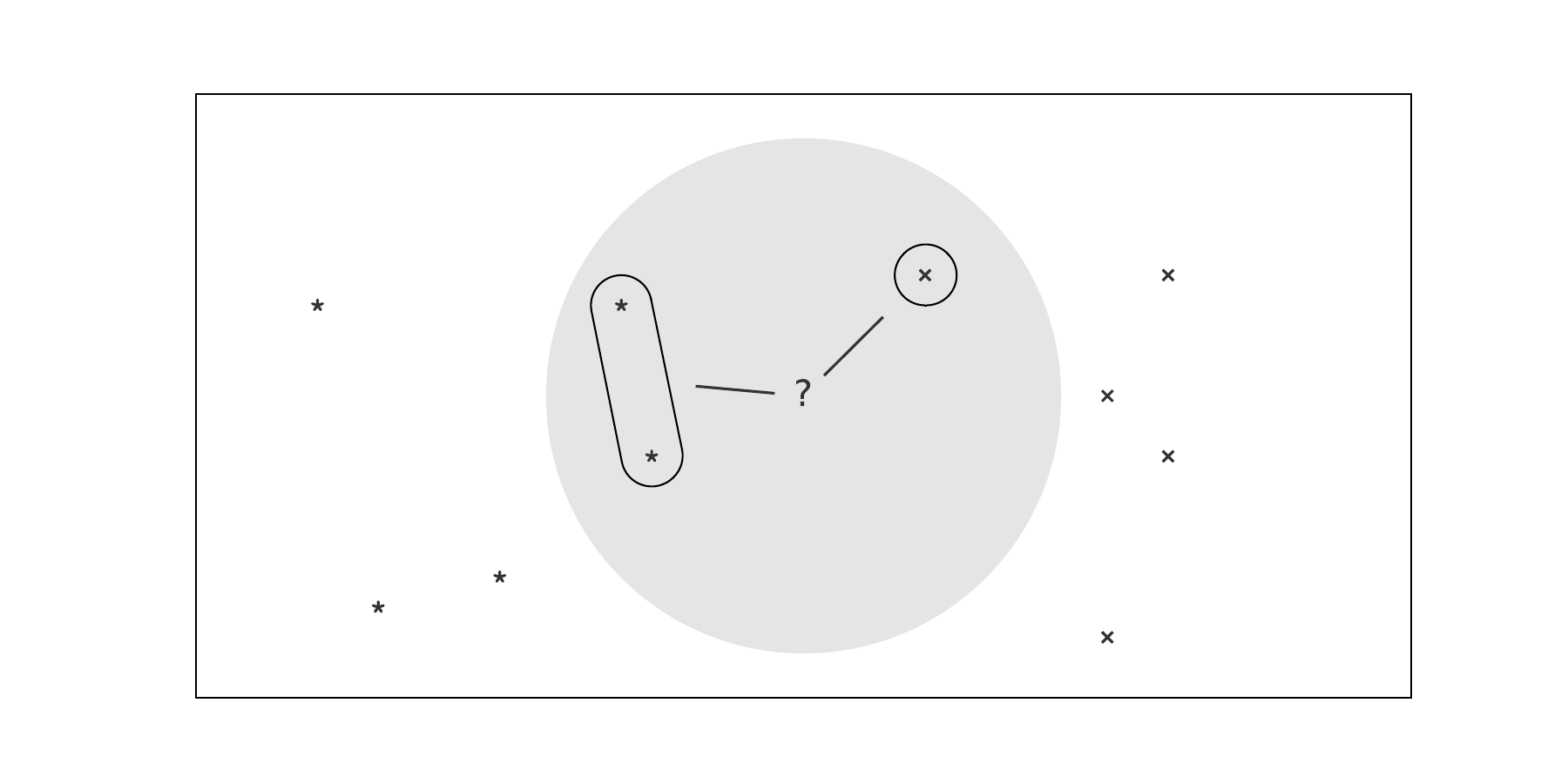}
\caption{Illustration of NN classification with $k = 3$. The class scores of a test record are calculated on the basis of its $k$ nearest training records.}
\label{fig_nn_example}
\end{figure}

In the early literature, nearest neighbour prediction (Fig.~\ref{fig_nn_example}) arose as a form of non-parametric (or \e{distribution-free}) statistical estimation, and was generally referred to as such. It was first formally presented in 1951 for classification, by Fix \& Hodges \cite{fix51discriminatory}. The idea to weigh the contribution of neighbours differently was initially proposed for regression, perhaps first by Watson \cite{watson64smooth}, Royall \cite{royall66class} and Shepard \cite{shepard68twodimensional}. This was inspired by an earlier idea to estimate the value of a density function in a point as a weighted sum, with weights corresponding inversely to the distances to the sample observations \cite{rosenblatt56remarks}. Dudani \cite{dudani73experimental,dudani76distance} appears to have been the first to propose weighted nearest neighbours for classification.

We can formally define weighted nearest neighbour classification as follows. Let $d$ be a distance measure and $k$ a positive integer, then the score for a decision class $C$ and a test record $y$ is:
 
 \begin{equation}
 \label{eq_wnn}
 \left.\sum_{i \leq k | x_i \in C} s_i \middle/ \sum_{i \leq k} s_i\right.,
 \end{equation}
where $x_i$ is the $i$th nearest neighbour of $y$ in the training set $X$ (according to $d$), and $s_i$ the weighted vote assigned to $x_i$, which remains to be defined. In practice, all proposals define $s_i$ in terms of the distance $d_i$ between $y$ and $x_i$ and/or the rank $i$. We will refer to these strategies as, respectively, distance- and rank-weighting. We recover classical unweighted nearest neighbour classification by choosing constant $s_i$, e.g. $s_i = 1$.

For rank-weighting, there have been proposals that let the weights depend linearly on the rank \cite{royall66class, dudani73experimental, dudani76distance, stone77consistent}, quadratically \cite{stone77consistent,altman92introduction}, reciprocally linearly \cite{gou11novel}, and according to the Fibonacci sequence \cite{pao07comparative}. Relatively recently, Samworth \cite{samworth12optimal} has established theoretically optimal weights:

 \begin{equation}
 \label{eq_samworth}
 s_i = \frac{1}{k}\left(1 + \frac{m}{2} - \frac{m}{2k^{\frac{2}{m}}}\left(i^{1 + \frac{2}{m}} - (i - 1)^{1 + \frac{2}{m}}\right)\right),
 \end{equation}
where $m$ is the dimensionality of the attribute space.

Proposals for distance-weighting have included weights that depend linearly on distance \cite{watson64smooth,dudani73experimental,dudani76distance}, reciprocally linearly \cite{dudani73experimental,dudani76distance} and reciprocally quadratically \cite{shepard68twodimensional}. Inspired by the work of Shepard \cite{shepard87toward}, Zavrel \cite{zavrel97empirical} has proposed Laplacian weights of the form $e^{-d_i}$.

The linear distance-weights given by Dudani \cite{dudani73experimental,dudani76distance} take the following form:

\begin{equation}
\label{eq_nn_weights_linear}
 s_i = \begin{dcases}
        \frac{d_k - d_i}{d_k - d_1} & k > 1;\\
        1 & k = 1.\\
       \end{dcases}
\end{equation}
For these weights, Dudani demonstrated a lower classification error than unweighted NN on a synthetic dataset. However, Bailey and Jain \cite{bailey78note} subsequently showed that this was due to the fact that Dudani had counted all ties as errors, and that when these are resolved instead (e.g. by randomly choosing a class), the performance of weighted and unweighted NN was similar on the synthetic dataset. Moreover, Bailey and Jain also proved that the asymptotic classification error of unweighted NN is minimal among all possible weighted variants of NN. This in turn elicited a response by Macleod et al. \cite{macleod87reexamination}, who argued that there exist finite classification problems where some distance-weighted variants of NN do have lower error. In order to demonstrate this, they used the following modified weights, which address the fact that in \eqref{eq_nn_weights_linear}, the $k$th weight is always 0 (if $k > 1$):

\begin{equation*}
\label{eq_nn_weights_macleod}
 s_i = \begin{dcases}
        \frac{d_k - d_i + d_k - d_1}{2(d_k - d_1)} & k > 1;\\
        1 & k = 1.\\
       \end{dcases}
\end{equation*}

Another modification of Dudani's linear weights was proposed by Gou et al. \cite{gou12new}:

\begin{equation*}
\label{eq_nn_weights_gou12}
 s_i = \frac{d_k - d_i}{d_k - d_1} \cdot \frac{d_k + d_1}{d_k + d_i}
\end{equation*}

Finally, Gou et al. \cite{gou11novel} have proposed a weighting scheme that combines the linear distance-weights of Dudani with reciprocally linear rank-weights:

\begin{equation*}
\label{eq_nn_weights_gou11}
 s_i = \frac{d_k - d_i}{d_k - d_1} \cdot \frac{1}{i}
\end{equation*}

\subsection{Kernel weighting}
\label{sec_kernel_weighting}

Nearest neighbour prediction is closely related to another form of non-parametric estimation, in which the prediction for a test record is a weighted sum of training values (e.g. \cite{priestley72nonparametric}). The weights are determined by a so-called kernel function that is applied to the distances between the test record and the training records. By choosing a decreasing kernel function, nearby training records receive greater weight, and by choosing a kernel function with finite support, the prediction is effectively limited to neighbours within a fixed distance, although the number of neighbours will generally vary for different test records.

There have been several proposals to extend the use of kernel functions to nearest neighbour-weighting. A very early proposal by Royall \cite{royall66class} applies a linear (or \e{triangular}) kernel to the scaled rank $\frac{i}{k}$, while Altman \cite{altman92introduction} suggests the use of a quadratic kernel. Both Wilson \& Martinez \cite{wilson97advances, wilson00integrated} and Hechenbichler \& Schliep \cite{hechenbichler04weighted} apply kernel functions to nearest neighbour distances, rescaled to values in $[0, 1]$ by dividing by the $k$th distance. In addition to constant and linear kernels, Wilson \& Martinez use Laplacian and Gaussian kernels, while Hechenbichler \& Schliep consider a biquadratic kernel $a \rma \frac{15}{16}(1 - a^2)^2$.

\subsection{Fuzzy nearest neighbour classification}
\label{sec_fnn}

\begin{figure}
\centering
\includegraphics[width=\linewidth]{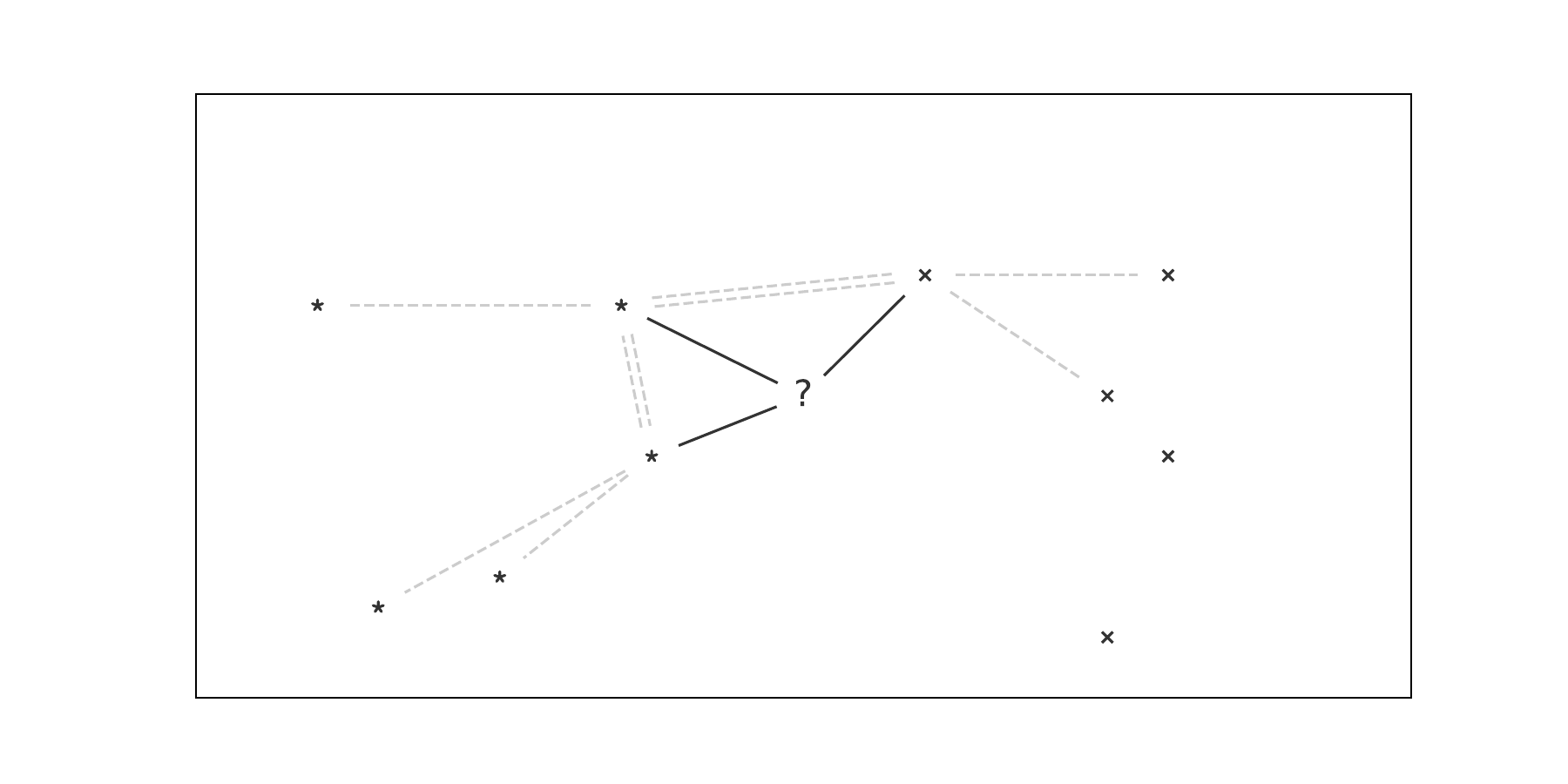}
\caption{Illustration of FNN classification with $k = 3$. The class scores of a test record are calculated on the basis of its $k$ nearest training records, whose class memberships are first fuzzified on the basis of their own $k$ nearest training records.}
\label{fig_fnn_example}
\end{figure}

There have been many proposals to modify nearest neighbour classification with fuzzy set theory \cite{derrac14fuzzy}. The most prominent of these is the fuzzy nearest neighbours (FNN) classifier (Fig.~\ref{fig_fnn_example}) of Keller et al. \cite{keller85fuzzy}. It defines the membership of a test record $y$ in the decision class $C$ as:

 \begin{equation}
 \label{eq_fnn}
 \left.\sum\limits_{i \leq k}u_i \cdot 1/d_i^{2/(q-1)}\middle/ \sum\limits_{i \leq k}1/d_i^{2/(q-1)}\right.,
 \end{equation}
for a choice of $q > 1$, where $d_i$ is the distance between $y$ and its $i$th nearest neighbour $x_i$, and $u_i$ is the class membership of $x_i$ in $C$. Keller et al. proposed two different options for $u_i$. Either $u_i$ can be chosen to be the crisp class membership of $x_i$ in $C$, or it can be fuzzified as follows:

 \begin{equation}
 \label{eq_fnn_ui}
 u_i = \begin{dcases}
        0.51 + 0.49\cdot n_C(x_i)/k & \text{if } x_i \in C;\\
        0.49\cdot n_C(x_i)/k & \text{if } x_i \nin C;\\
       \end{dcases},
 \end{equation}
where $n_C(x_i)$ is the number of neighbours of $x_i$ that belong to $C$, from among its $k$ nearest neighbours.

The original motivation for FNN given by Keller et al. was twofold. Firstly, by providing class scores \eqref{eq_fnn} rather than a crisp prediction that the test record $y$ belongs to one class or another, FNN expresses the `strength' of its class membership. Secondly, FNN lets the contribution of the nearest neighbours of $y$ depend both on their distance to $y$ and, when using the fuzzified class membership \eqref{eq_fnn_ui}, on how `typical' they are for a given decision class. However, the ability to provide class scores does not actually set it apart from classical NN \eqref{eq_wnn}. Likewise, we saw in Subsection~\ref{sec_nn_background} that there is a long tradition of letting the predictions of NN depend on the distance of the neighbours of $y$. In Subsection~\ref{sec_fnn2}, we will further analyse how FNN is related to NN.

\subsection{Fuzzy rough nearest neighbour classification}

\begin{figure}
\centering
\includegraphics[width=\linewidth]{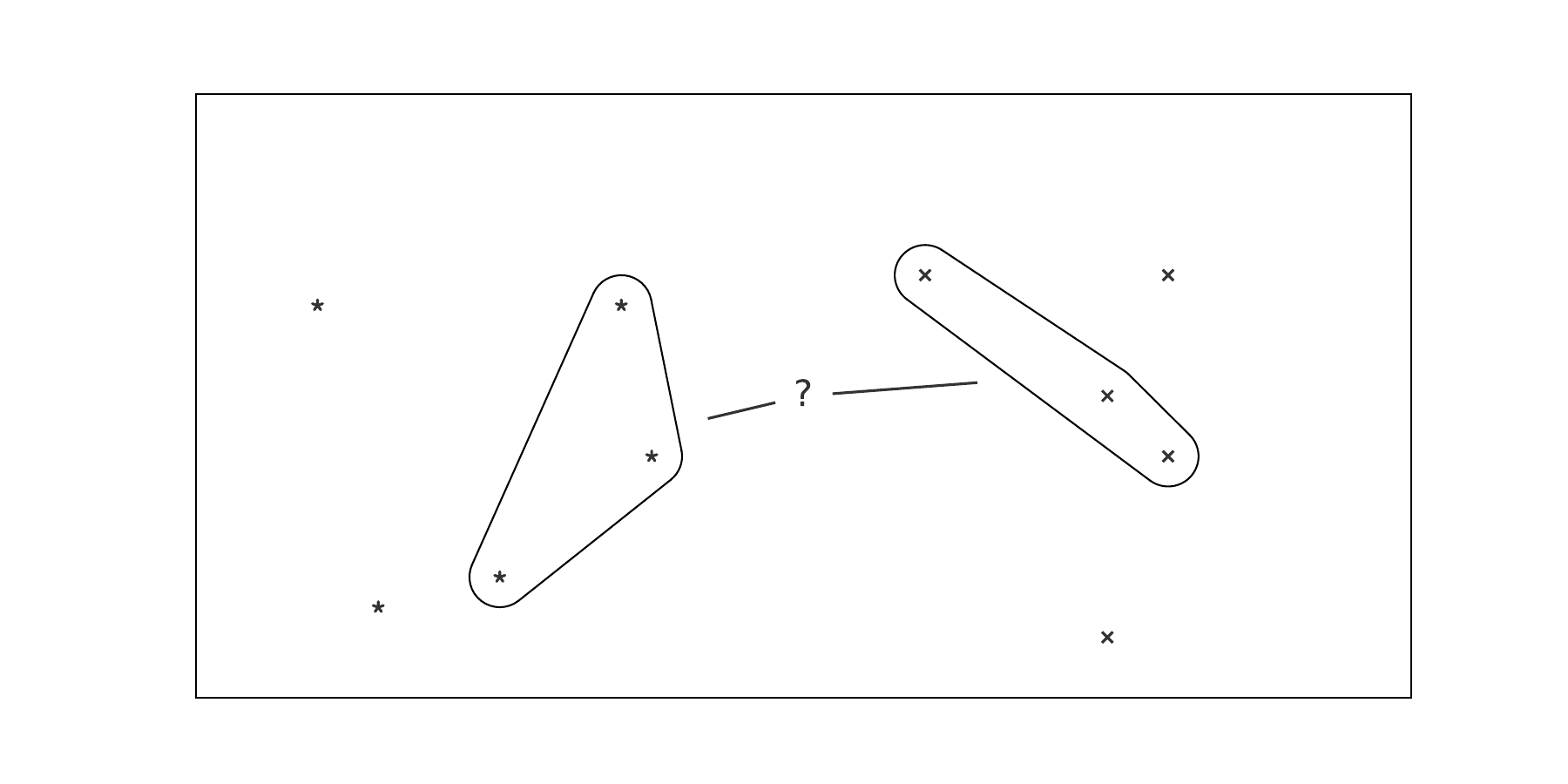}
\caption{Illustration of FRNN classification with $k = 3$. The class scores of a test record are calculated on the basis of its $k$ nearest training records within each decision class (upper approximation) as well as its $k$ nearest training records in the complement of each decision class (lower approximation).}
\label{fig_frnn_example}
\end{figure}

A more fundamentally different proposal has come in the form of fuzzy rough nearest neighbour (FRNN) classification (Fig.~\ref{fig_frnn_example}), originally proposed by Jensen \& Cornelis \cite{jensen08new}. This is based on fuzzy rough sets \cite{dubois90rough}, a fuzzified variant of rough sets \cite{pawlak81rough}. For each decision class $C$, we define two fuzzy sets, its upper approximation $\ol{C}$ and its lower approximation $\ul{C}$, as well as their mean, and the membership of a test record $y$ in any one of these can be used as a class score. A weighted variant of fuzzy rough sets was first introduced by Cornelis et al. \cite{cornelis10ordered}, and we use here the updated formulation of FRNN presented in \cite{lenz20scalable}. FRNN can be viewed as a form of nearest neighbour classification because the membership of $y$ in $\ol{C}$ and $\ul{C}$ is defined as a weighted sum of, respectively, its nearest neighbour similarities in $C$, and its nearest neighbour distances in $X\setminus C$.

Let $d$ be a distance measure and $k$ a positive integer, then the membership of a test record $y$ in $\ol{C}$ and $\ul{C}$ is defined, as, respectively:

 \begin{eqnarray*}
\begin{aligned}
\ol{C}(y) &:= \sum_{i \leq k} w_i \cdot (1 - d^+_i) / \sum_{i \leq k} w_i;\\
\ul{C}(y) &:= \sum_{i \leq k} w_i \cdot d^-_i / \sum_{i \leq k} w_i,
\end{aligned}
\end{eqnarray*}
where $d^+_i$ and $d^-_i$ are the $i$th nearest neighbour distance of $y$ in, respectively, $C$ and $X\setminus C$, and $w_i$ is a weight which depends on the rank $i$ and which remains to be defined.

Previous proposals for $w$ have included weight vectors that are constant \cite{cornelis10ordered} or that depend linearly \cite{verbiest12seleccion}, reciprocally \cite{verbiest14fuzzy} or exponentially \cite{cornelis10ordered} on the rank $i$ (see \cite{vluymans19weight} for an overview).

FRNN was originally proposed by Jensen \& Cornelis \cite{jensen08new} in reaction to a previous proposal by Sarkar \cite{sarkar07fuzzyrough} to modify FNN with a so-called fuzzy rough ownership function. In contrast to the proposal by Sarkar, FRNN is not a modification of FNN, but uses the core concepts of the upper and lower approximation from fuzzy rough set theory to obtain classification scores. However, a common motivation for both proposals is the observation (originally by Sarkar) that the class membership values produced by FNN automatically sum to one, and that it therefore cannot identify situations where we do not have enough information to make a reliable prediction for the test record $y$, i.e. when $y$ doesn't particularly resemble any of the training records. In such a situation, FRNN predicts low upper approximation values and high lower approximation values for \e{all} decision classes.

\subsection{Previous experiments}

Despite the extensive literature on NN classification, there have only been a small number of experimental evaluations of weights and distances.

Working with 18 synthetic and real-life datasets, Wettschereck \cite{wettschereck94study} found that reciprocally linear distance-weights clearly outperform unweighted NN for Euclidean distance, and that there is no clear difference between Euclidean and Boscovich distance. Zavrel \cite{zavrel97empirical}, using cosine distance, additionally considered linear and Laplacian weights, but only linear weights clearly outperformed unweighted NN. Hechenbichler \& Schliep \cite{hechenbichler04weighted} only evaluated linear and biquadratic weights, on a small number of datasets, without drawing any firm conclusions. However, we note that they appear to obtain generally better results for Boscovich than for Euclidean distance. Gou et al. \cite{gou12new}, using Euclidean distance, found that their proposal outperforms both unweighted NN and linear distance-weights on twelve real-life datasets. Finally, the weights proposed by Gou et al. \cite{gou12new} also came out on top, along with linear and reciprocally quadratic distance-weights, in the extensive evaluations of time series classification conducted by Geler et al. \cite{geler16comparison,geler20weighted}.

\begin{table}[!t]
\renewcommand{\arraystretch}{1.8}
\caption{Kernel functions $f$, with $a \in [0, 1]$, $\lambda \in (-1, \infty)$, $p \in (0, \infty)$, and $m$ the number of features.}
\label{tab_kernels}
\footnotesize
%\centerfloat
\begin{tabular}{p{.3\linewidth}p{.25\linewidth}p{.3\linewidth}}
\toprule
Name & $f(a)$ & Used in \\
\midrule
\multicolumn{3}{l}{\e{Fuzzy negations}}\\
\midrule
Linear & $1 - a$ & \cite{watson64smooth, royall66class, dudani73experimental, stone77consistent, wilson97advances, hechenbichler04weighted, verbiest12seleccion}\\
Quadratic & $1 - a^2$ & \cite{stone77consistent, altman92introduction}\\
Biquadratic & $(1 - a^2)^2$ & \cite{hechenbichler04weighted}\\
Samworth & $1 - a^{\frac{2}{m}}$ & \cite{samworth12optimal}\\
Sugeno-$\lambda$ & $\frac{1 - a}{1 + \lambda \cdot a}$ & \cite{gou12new}\\
Yager-$p$ & $(1 - a^{p})^{\frac{1}{p}}$ & \\
\midrule
\multicolumn{3}{l}{\e{Other proper kernels}}\\
\midrule
Constant & $1$ & \cite{fix51discriminatory, wilson97advances, hechenbichler04weighted, cornelis10ordered}\\
Laplace & $e^{-a}$ & \cite{zavrel97empirical,wilson97advances}\\
Gauss & $e^{-a^2/2}$ & \cite{wilson97advances}\\
\midrule
\multicolumn{3}{l}{\e{Improper kernels}}\\
\midrule
Reciprocally linear & $\frac{1}{a}$ & \cite{dudani73experimental, keller85fuzzy, gou11novel, verbiest14fuzzy}\\
Reciprocally quadratic & $\frac{1}{a^2}$ & \cite{shepard68twodimensional, keller85fuzzy}\\

\bottomrule
\end{tabular}
\end{table}

\section{Proposals}
\label{sec_proposals}

In this section, we will discuss our novel proposals. These include a universal framework for nearest neighbour weighting for both NN and FRNN, a new weight type inspired by fuzzy Yager negation, an analysis of FNN, and a characterisation of scaling measures that relates them to distance measures.

\subsection{Kernels}

\begin{figure}
\centering
\includegraphics[width=\linewidth]{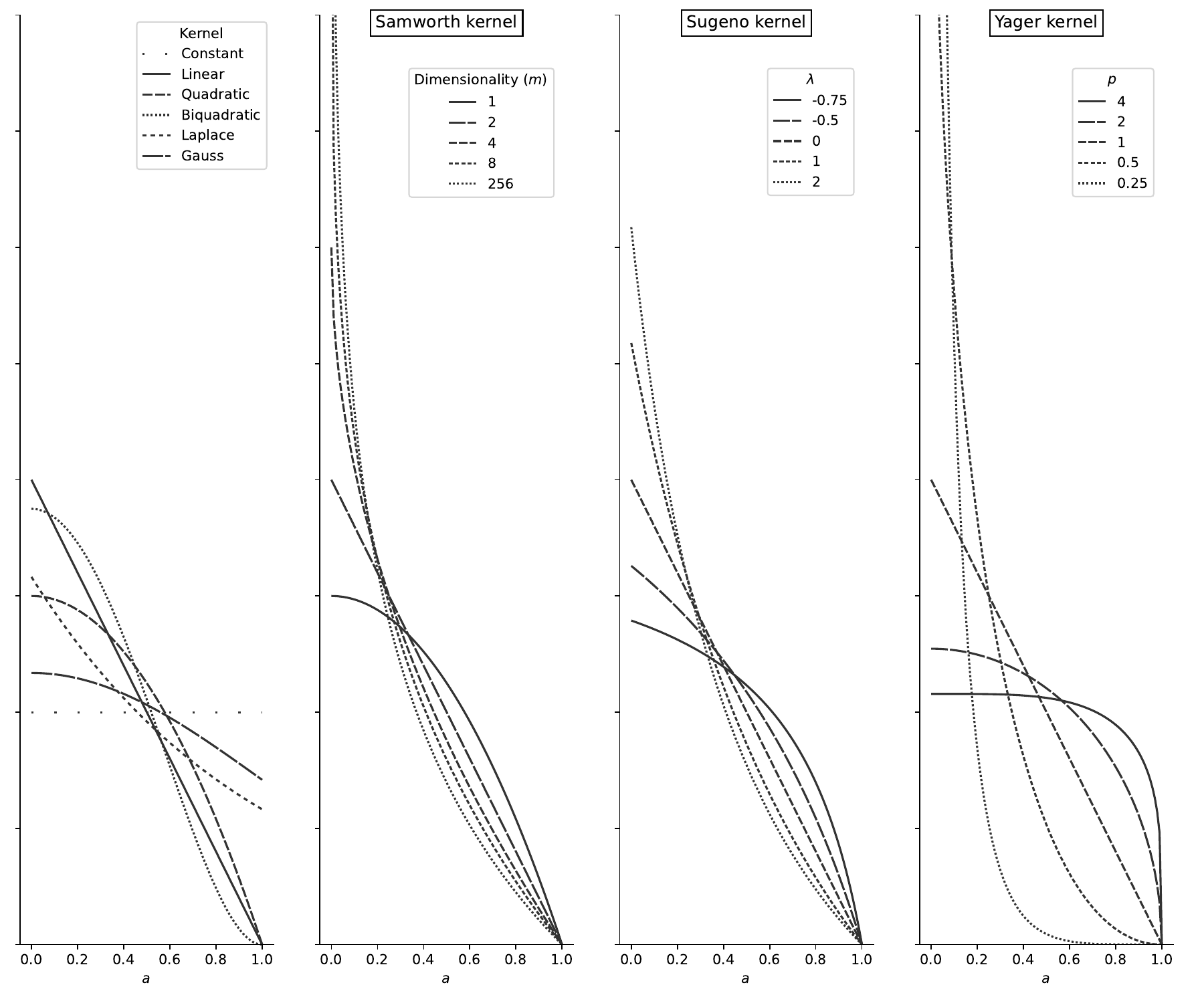}
\caption{Proper kernel functions, expressed in terms of $a \in [0, 1]$. In order to visualise the different weight that they place on smaller and larger values, we have rescaled each kernel by a constant, such that each kernel covers the same area on the interval $[0, 1]$. As the resulting absolute values are essentially arbitrary, we have deliberately left the vertical axis unmarked.}
\label{fig_kernels}
\end{figure}

In order to make nearest neighbour weighting easier to evaluate, we adopt a standardised representation based on kernel functions, inspired by the proposals discussed in Subsection~\ref{sec_kernel_weighting}.

The term \e{kernel} is used in various contexts for several related but distinct concepts. The type of kernel that we are interested in, from the statistical and signal processing literature, is a function on the real line that is symmetric around 0 as well as maximal around 0, such that it can be used to amplify values around 0 through multiplication. When such a kernel function has finite support, e.g. on $[-1, 1]$, it can be used more specifically as a window function to filter out all values outside this range.

In the context of nearest neighbour weighting, all ranks and distance values are non-negative, and the weights are distributed over a finite range, limited by the $k$th rank or nearest neighbour distance, which can be rescaled to the interval $[0, 1]$. Therefore, all weight types that we will consider can be covered by the following more restricted definition of a kernel:

\begin{definition}
 A (proper) \e{kernel} is a decreasing function $f: [0, 1] \lra \R_{\geq 0}$. A \e{normalised} kernel is a kernel with $f(0) = 1$. An \e{improper} kernel is a decreasing function $f: (0, 1] \lra \R_{\geq 0}$ with $\lim_{a \ra 0} f(a) = \infty$.
\end{definition}
Note that any proper kernel can be normalised through division by $f(0)$, such that $f(0) = 1$ and the range of $f$ becomes $[0, 1]$. An improper kernel is improper precisely because it cannot be normalised in this way.

A distinct concept from fuzzy set theory is that of a \e{fuzzy negation} (or \e{fuzzy complement}) \cite{higashi82measures}, which generalises the ordinary negation from classical logic that sends 0 to 1 and 1 to 0:

\begin{definition}
 A \e{fuzzy negation} is a decreasing function $f: [0, 1] \lra [0, 1]$ with $f(0) = 1$ and $f(1) = 0$.
\end{definition}

While these two concepts have different origins, we see that a fuzzy negation is in fact the same thing as a normalised kernel $f$ which satisfies $f(1) = 0$. Conversely, we can view proper and improper kernels as a loose form of fuzzy negation. This correspondence allows us to interpret the effect of applying a kernel not just as placing greater weight on the contribution of nearby neighbours, but also as converting distance values into similarity values.

In the following subsections, we will redefine NN, FNN and FRNN classification in terms of kernel functions, and show how the various weighting proposals from the literature can be expressed through kernels by rescaling the distance and rank values to the interval $[0, 1]$. The advantage of this approach is that, for most weight types, this eliminates the dependence of the weights on $k$, i.e. they can be expressed as a single kernel function that is `stretched' across the rank or distance values.\footnote{For linear weights, this is easy to see as the slope of the weights depends on $k$.} Thus, the kernel function captures the essence of a weighting type, and comparing these kernel functions reveals the essential differences between the various weighting proposals. The kernels discussed in this paper are listed in Table~\ref{tab_kernels} and the proper kernels are visualised in Fig.~\ref{fig_kernels}.

\subsection{NN}

Using the definition of a kernel function allows us to state the following generalised definition for weighted nearest neighbour classification:

\begin{definition}
Let $d$ be a distance measure, $k$ a positive integer, and $w$ and $s$ choices of kernel functions. Then the score for a decision class $C$ and a test record $y$ is:
 
 \begin{equation}
 \label{eq_nn_new}
\left.\sum_{i \leq k | x_i \in C} w(i^*) \cdot s(d_i^*) \middle/ \left(\sum_{i \leq k} w(i^*) \cdot s(d_i^*)\right)\right.,
 \end{equation}
where $x_i$ is the $i$th nearest neighbour of $y$ in the training set $X$ (as determined by $d$), $d_i$ the corresponding distance, $d_i^* := d_i/d_k$ and $i^* := \frac{i}{k + 1}$.

We adopt the following conventions to resolve specific edge cases:

\begin{itemize}
 \item If $d_k = 0$ (and therefore $d_i = 0$ for all $i \leq k$), we stipulate $d_i^* := d_i = 0$.
 \item If $d_1 = d_k$ (and therefore $d_i^* = 1$ for all $i \leq k$) and if $s(1) = 0$, we stipulate $s(d_i^*) := 1$ for all $i$.
 \item If $s$ is an improper kernel, and $d_i^* = 0$ for some $i$, we stipulate $s(d_i^*) := 1$ for all such $i$ and $s(d_i^*) := 0$ for all other $i$.
\end{itemize}
\end{definition}

When $w$ is constant, we recover NN with distance-weights, when $s$ is constant, we recover NN with rank-weights, and when both $w$ and $s$ are constant, we recover unweighted NN. In addition, in all three edge cases listed above, we also effectively revert to performing unweighted classification with (part of) the nearest neighbours of $y$.

In \ref{sec_proofs}, we show that most existing weighting proposals discussed in Subsection~\ref{sec_weighted_nearest_neighbour_classification} can be rewritten as kernel functions (Table~\ref{tab_kernels}, Fig.~\ref{fig_kernels}) applied to $i^*$ (rank-weighting) or $d_i^*$ (distance-weighting), making them compatible with our generalised definition of NN \eqref{eq_nn_new}. In particular, note that the linear (or \e{triangular}) kernel is the original form of fuzzy negation introduced by both {\L}ukasiewicz \cite{lukasiewicz23interpretacja} and Zadeh \cite{zadeh65fuzzy}, that the kernel corresponding to the weight type proposed by Gou et al. \cite{gou12new} is Sugeno-$\lambda$ negation with $\lambda = 1$, and that the rank-weights shown to be theoretically optimal by Samworth \cite{samworth12optimal} correspond to a kernel $a \rma 1 - a^{\frac{2}{m}}$, where $m$ is the dimensionality of the attribute space.

By reformulating both distance-weights and rank-weights in terms of a kernel function, we obtain a single unique way to characterise all the different weight types. In addition, this representation makes it clear that we could also choose to apply e.g. the Sugeno-1 kernel to obtain rank-weights, or the Samworth kernel to obtain distance-weights, even though they were originally proposed for, respectively, distance-weights and rank-weights. Furthermore, we could choose to apply both rank- and distance-weights at the same time, as in the proposal by Gou et al. \cite{gou11novel}, which can be realised by combining a reciprocally linear rank-kernel and a linear distance-kernel.

\subsection{Yager weights}

As mentioned in the previous subsections, kernels are slightly generalised fuzzy negation functions, and two fuzzy negations from the fuzzy logic literature correspond to existing weighting proposals: classical (linear) and Sugeno-$\lambda$ negation. There is in fact a third type of fuzzy negation that is frequently encountered in the literature, with the following form:

 \begin{equation*}
 \label{eq_yager}
a \lrma (1 - a^{p})^{\frac{1}{p}},
 \end{equation*}
for some $p > 0$. This is generally known as Yager negation, because it was proposed by Higashi \& Klir \cite{higashi82measures} to accompany other operators introduced by Yager \cite{yager80general}.

Just like classical and Sugeno negation, Yager negation can be used for nearest neighbour weighting. Specifically, we propose to use Yager negation with $p = \frac{1}{2}$, because the resulting contour is quite different from that of the existing weighting proposals, except the Samworth kernel for larger values of $m$ (Fig.~\ref{fig_kernels}).

\subsection{FNN}
\label{sec_fnn2}

Recall that in the original proposal of FNN (Subsection~\ref{sec_fnn}), there were two possible values for $u_i$. When $u_i$ is chosen crisply, \eqref{eq_fnn} simplifies to
 
 \begin{equation*}
 \label{eq_fnn_crisp}
 \left.\sum\limits_{i \leq k | x_i \in C}1/d_i^{2/(q-1)}\middle/ \sum\limits_{i \leq k}1/d_i^{2/(q-1)}\right..
 \end{equation*}
In other words, FNN becomes equivalent to NN classification \eqref{eq_wnn}, with

 \begin{equation*}
 \label{eq_fnn_weights}
 s_i = 1/d_j^{\frac{2}{q-1}},
 \end{equation*}
for some $q > 1$. When $q = 3$ and $q = 2$, we obtain, respectively, reciprocally linear and reciprocally quadratic distance-weights.

Therefore, we will continue with the fuzzy variant of $u_i$ \eqref{eq_fnn_ui}. In that case, \eqref{eq_fnn} becomes:

 \begin{equation*}
 \begin{aligned}
 \label{eq_fnn_fuzzy}
&\mathrel{\phantom{=}} \left.\sum\limits_{i \leq k}u_i \cdot s_i\middle/\sum\limits_{i \leq k}s_i\right.\\
&= \left.\left(\smashoperator[r]{\sum_{i \leq k | x_i \in C}}0.51 \cdot s_i + \sum\limits_{i \leq k}0.49\cdot n_C(x_i)/k \cdot s_i\right)\middle/\sum_{i \leq k}s_i\right.\\
&= \left.0.51 \cdot \sum_{\mathclap{i \leq k | x_i \in C}} s_i \middle/ \sum_{i \leq k} s_i\right. + \left.0.49 \cdot \sum_{i \leq k} n_C(x_i)/k \cdot s_i\middle/\sum_{i \leq k}s_i\right.\\
 \end{aligned}
 \end{equation*}
Thus, in this variant, the FNN class score is the weighted average of two components. The first component is, again, NN, while the second component is NN with fuzzified class membership. This second component can be rewritten as:

 \begin{equation*}
 \begin{aligned}
 \label{eq_fnn_approx}
&\mathrel{\phantom{=}} \left.\sum_{i \leq k} n_C(x_i)/k \cdot s_i\middle/\sum_{i \leq k}s_i\right.\\
&= \left.\frac{1}{k}\sum_{i,j \leq k | x_{ij} \in C} s_i\middle/\sum_{i \leq k}s_{i}\right.\\
 \end{aligned}
 \end{equation*}
where $x_{ij}$ is the $j$th neighbour of the $i$th neighbour of $y$. In effect, this is also NN, with class scores that are on the one hand diluted (being based not just on the class scores of the $k$ nearest neighbours of $y$, but also on the class scores of \e{their} $k$ nearest neighbours) and on the other hand concentrated (because nearby neighbours will more frequently appear as neighbours of neighbours). The net effect of this on classification performance remains to be evaluated experimentally, but it is hard to see how this reduces the influence of `atypical' training records, the original stated motivation for FNN.

\subsection{FRNN}
 
The upper and lower approximations of FRNN can similarly be rewritten using kernel functions:

\begin{definition}
Let $d$ be a distance measure and $k$ a positive integer, $w$ a choice of kernel function and $s$ a choice of fuzzy negation. Then the score for a decision class $C$ and a test record $y$ is:

 \begin{eqnarray*}
\begin{aligned}
\ol{C}(y) &:= \left.\sum_{i \leq k} w(i^*) \cdot s(\min(d_i^+/d_{*}^+, 1)) \middle/ \sum_{i \leq k} w(i^*)\right.;\\
\ul{C}(y) &:= \left.\sum_{i \leq k} w(i^*) \cdot (1 - s(\min(d_i^-/d_{*}^-, 1))) \middle/ \sum_{i \leq k} w(i^*)\right.,
\end{aligned}
\end{eqnarray*}
where $i^* := \frac{i}{k + 1}$, where $d^+_i$ and $d^-_i$ are the $i$th nearest neighbour distance of $y$ in, respectively, $C$ and $X\setminus C$, and where $d^+_{*}$ and $d^-_{*}$ are to be defined.
\end{definition}
$d_{*}^+$ and $d_{*}^-$ determine cutoff values --- all larger distances are mapped to the minimum degree of similarity, typically 0. These cutoff values have to be constant across decision classes and test records, to allow for a proper comparison of class scores. If we choose values that are too small, $\min(d_i^+/d_{*}^+, 1)$ and $\min(d_i^-/d_{*}^-, 1)$ become equal to 1 for many test records and many values of $i$, and we lose information. If we choose values that are too large, $\min(d_i^+/d_{*}^+, 1)$ and $\min(d_i^-/d_{*}^-, 1)$ are generally close to $0$, and we do not make full use of the profile of the kernel $s$. Therefore, as a compromise, we calculate $d_k^+$ and $d_k^-$ of all training records, for all decision classes, and take $d_{*}^+$ and $d_{*}^-$ to be the respective maxima of these values.

Note that unlike NN, FRNN cannot be used with constant distance-weights, because this would equalise all class scores. Instead, the default choice is linear distance-weights, in which case the double negation $(1 - s(d_i))$ in the lower approximation simplifies to $d_i$. In addition, the exponentially decreasing rank-weights that have occasionally been proposed in the literature have a very limited usefulness, as the contribution of each additional value quickly becomes insignificant, and, eventually, impossible to compute.

\subsection{Distance and scaling measures}
\label{sec_minkowski_size}

Three distance measures that are frequently used with nearest neighbour classification are Euclidean, Boscovich (or \e{city-block}) and Chebyshev (or \e{maximum}) distance. These can be viewed, respectively, as the special cases $p = 2$, $p = 1$ and $p \ra \infty$ of the Minkowski $p$-distance between two points $x, y \in \R^m$ (for some $m \geq 1$):

\begin{equation*}
 \left|y - x\right|_p := \left(\sum_{i \leq m} \left\vert y_i - x_i \right \vert^p\right)^{\frac{1}{p}}.
\end{equation*}
The parameter $p$ determines the contribution of the per-attribute differences $y_i - x_i$ to the distance value. When $p = 1$ (Boscovich distance), the contribution of all per-attribute differences is equal, whereas when $p \ra \infty$ (Chebyshev distance), the distance is simply equal to the largest per-attribute difference, and all smaller per-attribute differences make no contribution. This means that with Chebyshev distance, two records are only neighbours if \e{all} their attribute values are similar, whereas with Boscovich distance, it is sufficient that their attribute values are similar \e{on average}.

The case $p = 2$ (Euclidean distance) stakes a middle ground between these two extremes, in that smaller per-attribute differences make less than a proportional contribution to the distance, without being completely ignored. A unique property of Euclidean distance is that it is invariant under rotations of the attribute space. Therefore, Euclidean distance is the appropriate distance measure when applying a rotation would result in an equivalent representation of the data.

In order to obtain a comparable contribution from all attributes, these must be rescaled to a common scale. This can be done by taking a measure of dispersion, and dividing each attribute by this measure, such that it becomes 1 for each attribute. Two common choices are the standard deviation and the half-range of each attribute. These can be linked to the concept of Minkowski $p$-distance by defining the Minkowski $p$-centre and $p$-radius of a dataset:

\begin{definition}
\label{def_p_radius}
Let $X = (x_1, x_2, \dots, x_n)$ be a univariate real-valued dataset. The Minkowski $p$-radius $r_p$ of $X$ is defined as:

\begin{equation*}
\begin{aligned}
 r_p(X) := \min_{z \in \R} \left(\frac{1}{n}\sum_{i \leq n} \left\vert x_i - z \right \vert^p\right)^{\frac{1}{p}},\\
\end{aligned}
 \end{equation*}
while the Minkowski $p$-centre of $X$ is the corresponding minimising value for $z$ (not necessarily unique for $p \leq 1$).
\end{definition}

The standard deviation and half-range of a dataset are $r_2$ and $r_{\infty}$, while the corresponding $2$-centre and $\infty$-centre of a dataset are its mean and its midrange. The $1$-centre of a dataset is its median, and the corresponding measure of dispersion $r_1$ that it minimises is the mean absolute deviation around the median. Thus, $r_1$ is another measure of dispersion that we can use to scale attributes with.

A potential advantage of $r_1$-scaling over $r_2$-scaling is its reduced sensitivity to outliers, as $r_1$ only depends linearly on outliers, rather than quadratically like $r_2$. In turn, both measures are much less sensitive to outliers than $r_{\infty}$, which is completely determined by the most extreme outlier. An alternative way to obtain a measure of dispersion that is less sensitive to outliers is to explicitly ignore peripheral values. Specifically, we will consider the semi-interquartile range $r_{\infty}^*$, which is the half-range of the central 50\% of all values.

\begin{table*}
\caption{Real-life Classification Datasets from the UCI Repository for Machine Learning. $n$: number of records; $c$: number of classes; $m$: number of attributes; IR: imbalance ratio.}
\label{tab_statistics}
\begin{tabular}{lrrrrlrrrr}
\toprule
Dataset & $n$ & $c$ & $m$ & IR & Dataset & $n$ & $c$ & $m$ & IR \\
\midrule
accent & 329 & 6 & 12 & 2.5 & mfeat & 2000 & 10 & 649 & 1.0 \\
acoustic-features & 400 & 4 & 50 & 1.0 & miniboone & 130\,064 & 2 & 50 & 2.6 \\
ai4i2020 & 10\,000 & 2 & 6 & 28.5 & new-thyroid & 215 & 3 & 5 & 3.5 \\
alcohol & 125 & 5 & 12 & 1.0 & oral-toxicity & 8992 & 2 & 1024 & 11.1 \\
androgen-receptor & 1687 & 2 & 1024 & 7.5 & page-blocks & 5473 & 5 & 10 & 31.6 \\
avila & 20\,867 & 12 & 10 & 38.7 & phishing-websites & 11\,055 & 2 & 30 & 1.3 \\
banknote & 1372 & 2 & 4 & 1.2 & plrx & 182 & 2 & 12 & 2.5 \\
bioaccumulation & 779 & 3 & 9 & 4.3 & pop-failures & 540 & 2 & 18 & 10.7 \\
biodeg & 1055 & 2 & 41 & 2.0 & post-operative & 87 & 2 & 8 & 2.6 \\
breasttissue & 106 & 6 & 9 & 1.3 & qualitative-bankruptcy & 250 & 2 & 6 & 1.3 \\
ca-cervix & 72 & 2 & 19 & 2.4 & raisin & 900 & 2 & 7 & 1.0 \\
caesarian & 80 & 2 & 5 & 1.4 & rejafada & 1996 & 2 & 6824 & 1.0 \\
ceramic & 37 & 4 & 34 & 1.4 & rice & 3810 & 2 & 7 & 1.3 \\
cmc & 1473 & 3 & 9 & 1.6 & seeds & 210 & 3 & 7 & 1.0 \\
codon-usage & 13\,011 & 20 & 64 & 25.5 & segment & 2310 & 7 & 19 & 1.0 \\
coimbra & 116 & 2 & 9 & 1.2 & seismic-bumps & 2584 & 2 & 18 & 14.2 \\
column & 310 & 3 & 6 & 1.9 & sensorless & 58\,509 & 11 & 48 & 1.0 \\
debrecen & 1151 & 2 & 19 & 1.1 & sepsis-survival & 110\,204 & 2 & 3 & 12.6 \\
dermatology & 358 & 6 & 34 & 2.2 & shuttle & 58\,000 & 7 & 9 & 560.8 \\
diabetes-risk & 520 & 2 & 16 & 1.6 & skin & 245\,057 & 2 & 3 & 3.8 \\
divorce & 170 & 2 & 54 & 1.0 & somerville & 143 & 2 & 6 & 1.2 \\
dry-bean & 13\,611 & 7 & 16 & 2.3 & sonar & 208 & 2 & 60 & 1.1 \\
ecoli & 332 & 6 & 7 & 6.3 & south-german-credit & 1000 & 2 & 20 & 2.3 \\
electrical-grid & 10\,000 & 2 & 12 & 1.8 & spambase & 4601 & 2 & 57 & 1.5 \\
faults & 1941 & 7 & 27 & 3.9 & spectf & 267 & 2 & 44 & 3.9 \\
fertility & 100 & 2 & 9 & 7.3 & sportsarticles & 1000 & 2 & 59 & 1.7 \\
flowmeters & 361 & 4 & 44 & 1.7 & sta-dyn-lab & 6248 & 2 & 244 & 9.5 \\
forest-types & 523 & 4 & 9 & 1.8 & tcga-pancan-hiseq & 801 & 5 & 20\,531 & 1.9 \\
gender-gap & 3145 & 2 & 15 & 7.9 & thoraric-surgery & 470 & 2 & 16 & 5.7 \\
glass & 214 & 6 & 9 & 3.6 & transfusion & 748 & 2 & 4 & 3.2 \\
haberman & 306 & 2 & 3 & 2.8 & tuandromd & 4464 & 2 & 241 & 4.0 \\
hcv & 589 & 2 & 12 & 9.5 & urban-land-cover & 675 & 9 & 147 & 2.2 \\
heart-failure & 299 & 2 & 12 & 2.1 & vehicle & 846 & 4 & 18 & 1.1 \\
house-votes-84 & 435 & 2 & 16 & 1.6 & warts & 180 & 2 & 8 & 2.0 \\
htru2 & 17\,898 & 2 & 8 & 9.9 & waveform & 5000 & 3 & 21 & 1.0 \\
ilpd & 579 & 2 & 10 & 2.5 & wdbc & 569 & 2 & 30 & 1.7 \\
ionosphere & 351 & 2 & 34 & 1.8 & wifi & 2000 & 4 & 7 & 1.0 \\
iris & 150 & 3 & 4 & 1.0 & wilt & 4839 & 2 & 5 & 17.5 \\
landsat & 6435 & 6 & 36 & 1.7 & wine & 178 & 3 & 13 & 1.3 \\
leaf & 340 & 30 & 14 & 1.2 & wisconsin & 683 & 2 & 9 & 1.9 \\
letter & 20\,000 & 26 & 16 & 1.0 & wpbc & 138 & 2 & 32 & 3.9 \\
lrs & 527 & 7 & 100 & 12.6 & yeast & 1484 & 10 & 8 & 11.6 \\
magic & 19\,020 & 2 & 10 & 1.8 &  &  &  &  &  \\
\bottomrule
\end{tabular}
\end{table*}

\section{Experimental setup}
\label{sec_nn_experimental_setup}

To evaluate NN, FNN and FRNN classification, we will use 85 numerical real-life datasets from the UCI repository for machine learning (Table~\ref{tab_statistics}). We perform 5-fold cross-validation, and calculate the mean area under the receiver operating characteristic (AUROC) as a measure of the discriminative ability of each classifier. For FRNN, this requires normalising class scores such that they sum to 1. To compare two alternatives, we calculate the $p$-value from a one-sided Wilcoxon signed-rank test. Where appropriate, we also apply the Benjamini-Hochberg method \cite{benjamini95controlling} to control the false discovery rate.

For all of NN, FNN and FRNN, we optimise $k$ through leave-one-out validation, evaluating all values up to a certain value $k^{\max}$. For FRNN, we also choose between the upper, lower or mean approximation based on validation AUROC.

Determining the best validation procedure for hyperparameter selection is a surprisingly complex issue. As estimators of the test set score, validation procedures suffer from both bias and variance. Compared to e.g. 10-fold cross-validation, leave-one-out validation has lower bias, but can have higher variance when the model is unstable \cite{zhang15crossvalidation}, which depends on the classification procedure, the evaluation measure and the structure of the dataset, in particular the amount of outliers \cite{bengio04no}. Complicating things further, a validation procedure that is worse for estimating prediction performance can still, somewhat counter-intuitively, be better for selecting the optimal hyperparameter values, and vice-versa \cite{zhang15crossvalidation}.

In our case, there are three particular reasons that justify the use of leave-one-out validation. Firstly, nearest neighbour classification is very stable \cite{bousquet02stability}. Secondly, the optimal value of the hyperparameter that we optimise ($k$) is likely influenced by the dataset size, so it is advantageous that the internal training set size of leave-one-out validation ($n-1$) is nearly the same as that of the full training set ($n$). And thirdly, leave-one-out validation can be performed very efficiently for nearest neighbour algorithms, by executing a single $(k+1)$-nearest neighbour query on the training set and deleting every match between a training record and itself.

In addition to the use of leave-one-out validation, there is a second reason why $k$ can be optimised very efficiently: we do not need to perform a new nearest neighbour query for every value of $k$ that we want to evaluate. Instead, we only need to perform a single $k^{\max}$-nearest neighbour query, after which we can evaluate all values of $k$ from $k^{\max}$ to 1 by iteratively deleting all $(k + 1)$-nearest neighbour distances.

Because larger datasets may require larger values of $k$, we let $k^{\max}$ depend logarithmically on the training set size $n$. In order to ensure high model quality, we set $k^{\max} = 100 \log n$. For our largest dataset, \e{skin}, this means that $k^{\max} = 1219$ in each fold.

\section{Results}
\label{sec_nn_results}

In this section, we will present the results of our experiments. To start with, we will evaluate distance measures, scaling measures and weight types, but restrict ourselves to the weight types that have previously been proposed in the literature. We will then ask whether these results can be further improved upon by using the Yager-$\frac{1}{2}$ weights that we have proposed.

\begin{table}
\centering
\caption{One-sided $p$-values, Samworth distance-weights vs other distance-weights, for NN with Boscovich distance, in terms of AUROC. With Benjamini-Hochberg false discovery rate correction in each column.}
\label{tab_p_values_nn_kernels}
\begin{tabular}{lllll}
\toprule
Samworth vs\dots & \multicolumn{4}{l}{Scaling} \\
 & $r_1$ & $r_2$ & $r_{\infty}$ & $r_{\infty}^*$ \\
\midrule
Constant & $< 0.0001$ & $< 0.0001$ & $< 0.0001$ & $< 0.0001$ \\
Quadratic & $0.00017$ & $< 0.0001$ & $< 0.0001$ & $< 0.0001$ \\
Gauss & $< 0.0001$ & $< 0.0001$ & $< 0.0001$ & $< 0.0001$ \\
Laplace & $< 0.0001$ & $< 0.0001$ & $< 0.0001$ & $< 0.0001$ \\
Linear & $< 0.0001$ & $0.00011$ & $0.00017$ & $< 0.0001$ \\
Macleod & $< 0.0001$ & $< 0.0001$ & $< 0.0001$ & $< 0.0001$ \\
Biquadratic & $0.0075$ & $0.11$ & $0.71$ & $0.024$ \\
Reciprocally linear & $0.00044$ & $< 0.0001$ & $< 0.0001$ & $< 0.0001$ \\
Reciprocally quadratic & $0.038$ & $0.011$ & $0.025$ & $0.059$ \\
Sugeno-1 & $0.0029$ & $0.0015$ & $0.012$ & $0.00015$ \\
\bottomrule
\end{tabular}
\end{table}

\subsection{NN}
\label{sec_results_nn}

We first consider the effect of the distance on classification performance. We find that for all types of scaling and all weight types, Boscovich distance leads to significantly better performance than Euclidean ($p < 0.0031$) and Chebyshev ($p < 2.0 \cdot 10^{-9}$) distance. For this reason, we will continue with Boscovich distance only. However, we investigate this question in greater detail in Subsection~\ref{sec_results_distance_measures}.

Next, we have a look at the different kernels. First we compare using each kernel for distance-weights versus rank-weights, for each type of scaling. With three exceptions, we find that distance-weights lead to significantly better classification performance than rank-weights ($p < 0.0019$). The exceptions are the reciprocally linear and Laplace kernels, for which the difference is not or only weakly significant and to which we return below, as well as the Gaussian kernel, for which the difference is only weakly significant with $r_{\infty}$-scaling ($p = 0.26$).

Among the distance-weights, the Samworth kernel significantly outperforms all other kernels (Table~\ref{tab_p_values_nn_kernels}), with three exceptions. The difference with respect to the biquadratic and reciprocally quadratic kernel is only weakly significant with, respectively, $r_2$- and $r_{\infty}^*$-scaling. Moreover, with $r_{\infty}$-scaling, the biquadratic kernel is actually slightly better than the Samworth kernel on our data, but we will see below that $r_{\infty}$-scaling is suboptimal. Finally, we also find that Samworth distance-weights outperform reciprocally linear ($p < 0.0048$) and Laplace ($p < 6.2 \cdot 10^{-6}$) rank-weights across scaling measures.

Samworth distance-weights also significantly outperform the combination of linear distance-weights and reciprocal rank-weights proposed by Gou et al. \cite{gou11novel} for all scaling types ($p < 0.044$) except $r_{\infty}$, where the difference is only weakly significant ($p = 0.13$). However, our general formula for NN classification also allows for other combinations. Indeed, we find that Samworth distance-weights are outperformed by the combination of Samworth distance-weights and Samworth rank-weights ($p < 0.040$).

\begin{table}
\centering
\caption{One-sided $p$-values, various scalings vs $r_{\infty}$-scaling, for NN with Boscovich distance and Samworth distance- and rank-weights, in terms of AUROC. With Benjamini-Hochberg false discovery rate correction.}
\label{tab_p_values_scales}
\begin{tabular}{ll}
\toprule
Test & p \\
\midrule
$r_{\infty}^*$ vs $r_{\infty}$ & $0.013$ \\
$r_1$ vs $r_{\infty}$ & $0.013$ \\
$r_2$ vs $r_{\infty}$ & $0.00044$ \\
\bottomrule
\end{tabular}
\end{table}

Finally, when we consider the different measures of dispersion that can be used to normalise a dataset through rescaling, we find that $r_1$, $r_2$ and $r_{\infty}^*$ do not significantly outperform each other for the combination of Samworth distance- and rank-weights, while they all outperform $r_{\infty}$ (Table~\ref{tab_p_values_scales}). For other weight types, we obtain comparable results.

\subsection{FNN}
\label{sec_results_fnn}

For FNN, we consider reciprocally linear and reciprocally quadratic distance-weights, as well as Samworth distance-weights and a combination of Samworth rank- and distance-weights, since we found in the previous Subsection that these latter two perform well for classical NN.

As with NN, FNN performs significantly better with Boscovich distance than with either Euclidean ($p < 0.025$) or Chebyshev ($p < 8.3 \cdot 10^{-7}$) distance.

\begin{table}
\centering
\caption{One-sided $p$-values, Samworth distance-weights vs other distance-weights, for FNN with Boscovich distance, in terms of AUROC. With Benjamini-Hochberg false discovery rate correction in each column.}
\label{tab_p_values_fnn_distance_kernels}
\begin{tabular}{lllll}
\toprule
Samworth vs\dots & \multicolumn{4}{l}{Scaling} \\
 & $r_1$ & $r_2$ & $r_{\infty}$ & $r_{\infty}^*$ \\
\midrule
Reciprocally linear & $0.028$ & $0.17$ & $0.012$ & $0.68$ \\
Reciprocally quadratic & $0.34$ & $0.32$ & $0.18$ & $0.69$ \\
\bottomrule
\end{tabular}
\end{table}

\begin{table}
\centering
\caption{One-sided $p$-values, comparing the scaler in each row against the scaler in each column, for FNN with Boscovich distance and Samworth distance-weights, in terms of AUROC. With Benjamini-Hochberg false discovery rate correction in each row.}
\label{tab_p_values_fnn_scales}
\begin{tabular}{llll}
\toprule
 & $r_2$ & $r_{\infty}$ & $r_{\infty}^*$ \\
\midrule
$r_1$ & $0.19$ & $0.076$ & $0.076$ \\
$r_2$ &  & $0.15$ & $0.15$ \\
$r_{\infty}$ &  &  & $0.98$ \\
\bottomrule
\end{tabular}
\end{table}

\begin{table}
\centering
\caption{One-sided $p$-values, Samworth vs linear distance-weights, for FRNN with Boscovich distance and various rank-weights and scalings.}
\label{tab_p_values_frnn_distance_kernels}
\begin{tabular}{lllll}
\toprule
Rank-kernel & \multicolumn{4}{l}{Scaling} \\
 & $r_1$ & $r_2$ & $r_{\infty}$ & $r_{\infty}^*$ \\
\midrule
Constant & $< 0.0001$ & $< 0.0001$ & $< 0.0001$ & $< 0.0001$ \\
Linear & $< 0.0001$ & $< 0.0001$ & $< 0.0001$ & $< 0.0001$ \\
Reciprocally linear & $0.082$ & $0.12$ & $0.061$ & $0.0061$ \\
Samworth & $0.0012$ & $0.00019$ & $0.00043$ & $< 0.0001$ \\
\bottomrule
\end{tabular}
\end{table}

\begin{table}
\centering
\caption{One-sided $p$-values, Samworth rank-weights vs other rank-weights, for FRNN with Boscovich distance and Samworth distance-weights. Holm-Bonferioni correction applied in each column.}
\label{tab_p_values_frnn_rank_kernels}
\begin{tabular}{lllll}
\toprule
Samworth vs\dots & \multicolumn{4}{l}{Scaling} \\
 & $r_1$ & $r_2$ & $r_{\infty}$ & $r_{\infty}^*$ \\
\midrule
Constant & $0.080$ & $0.0033$ & $0.089$ & $0.0048$ \\
Linear & $0.080$ & $0.074$ & $0.73$ & $0.32$ \\
Reciprocally linear & $0.080$ & $0.016$ & $0.11$ & $0.043$ \\
\bottomrule
\end{tabular}
\end{table}

However, it is not as clear as with NN that Samworth distance-weights perform better than reciprocally linear or reciprocally quadratic weights (Table~\ref{tab_p_values_fnn_distance_kernels}). Furthermore, the combination of Samworth rank- and distance-weights actually performs worse than Samworth distance-weights alone for $r_1$ ($p = 0.032$) and $r_{\infty}$ ($p = 0.000078$) scaling, while for $r_2$- and $r_{\infty}^*$-scaling, the difference is not significant.
We have weak evidence that with FNN, $r_1$-scaling leads to better performance than $r_2$-scaling, and that in turn both are preferable over $r_{\infty}$- and $r_{\infty}^*$-scaling (Table~\ref{tab_p_values_fnn_scales}).

\subsection{FRNN}
\label{sec_results_frnn}

\begin{table}
\centering
\caption{One-sided $p$-values, various scalings vs $r_{\infty}$-scaling, for FRNN with Boscovich distance and Samworth distance- and rank-weights, in terms of AUROC. With Benjamini-Hochberg false discovery rate correction.}
\label{tab_p_values_frnn_scales}
\begin{tabular}{ll}
\toprule
Test & p \\
\midrule
$r_{\infty}^*$ vs $r_{\infty}$ & $0.033$ \\
$r_1$ vs $r_{\infty}$ & $0.033$ \\
$r_2$ vs $r_{\infty}$ & $0.0078$ \\
\bottomrule
\end{tabular}
\end{table}

For FRNN, we evaluate the different types of rank-weights proposed in the literature, corresponding to the constant, linear and reciprocally linear kernel, in combination with linear distance-weights. In addition, we evaluate Samworth rank- and distance-weights, motivated by their excellent performance with NN.

As with NN and FNN, we find that Boscovich distance leads to higher AUROC than Euclidean ($p < 0.00085$) and Chebyshev ($p < 6.2 \cdot 10^{-10}$) distance for all combinations of kernels and scaling measures.

The traditional choice for distance-weights is to use a linear kernel, but we find that the Samworth kernel performs better, although the difference is only weakly significant in combination with reciprocally linear rank-weights (Table~\ref{tab_p_values_frnn_distance_kernels}). Likewise, Samworth rank-weights appear to be the best choice, but the advantage over other kernels is only weakly significant (Table~\ref{tab_p_values_frnn_rank_kernels}).

As with NN, the measures of dispersion $r_1$, $r_2$ and $r_{\infty}^*$ do not significantly outperform each other, but do outperform $r_{\infty}$, although even this latter fact is only weakly significant (Table~\ref{tab_p_values_frnn_scales}).

\subsection{NN vs FNN vs FRNN}

\begin{table}
\centering
\caption{One-sided $p$-values, NN vs FNN with Boscovich distance and various distance kernels and scalings, in terms of AUROC.}
\label{tab_p_values_nn_vs_fnn}
\begin{tabular}{lllll}
\toprule
Distance-kernel & \multicolumn{4}{l}{Scaling} \\
 & $r_1$ & $r_2$ & $r_{\infty}$ & $r_{\infty}^*$ \\
\midrule
Reciprocally linear & $< 0.0001$ & $< 0.0001$ & $< 0.0001$ & $< 0.0001$ \\
Reciprocally quadratic & $< 0.0001$ & $< 0.0001$ & $< 0.0001$ & $< 0.0001$ \\
Samworth & $< 0.0001$ & $< 0.0001$ & $< 0.0001$ & $< 0.0001$ \\
\bottomrule
\end{tabular}
\end{table}

\begin{table}
\centering
\caption{One-sided $p$-values, FRNN vs NN, Boscovich distance and Samworth rank- and distance-weights.}
\label{tab_p_values_frnn_vs_nn}
\begin{tabular}{llll}
\toprule
\multicolumn{4}{l}{Scaling} \\
$r_1$ & $r_2$ & $r_{\infty}$ & $r_{\infty}^*$ \\
\midrule
$0.021$ & $0.0092$ & $0.0052$ & $0.049$ \\
\bottomrule
\end{tabular}
\end{table}

\begin{figure}
\centering
\includegraphics[width=\linewidth]{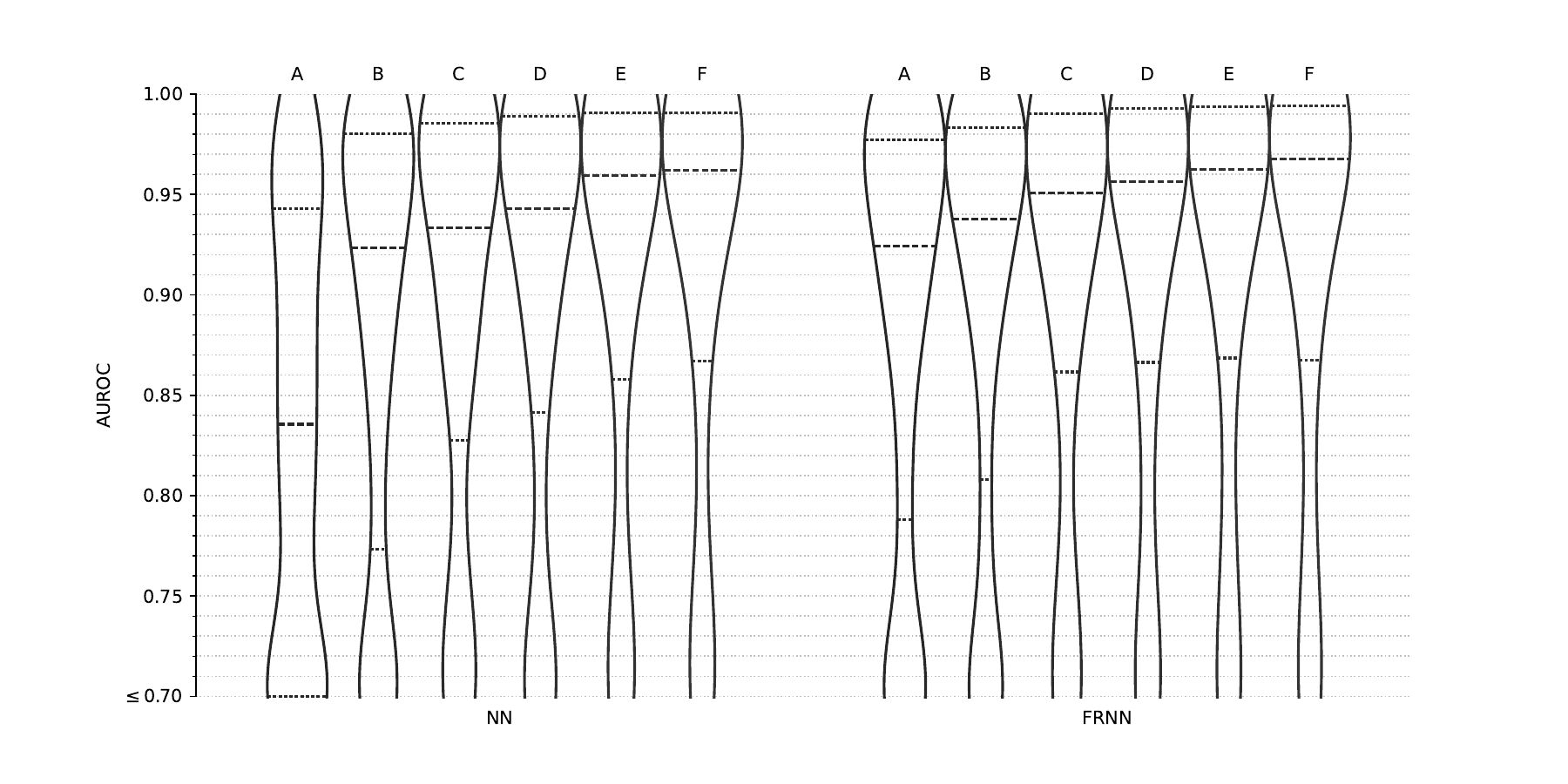}
\caption{Violin plots of the distribution of AUROC scores across our selection of datasets, illustrating the cumulative benefit of choosing optimal hyperparameter values. We start with \textbf{A}: choosing $k = 1$, Euclidean distance and $r_{\infty}$-scaling, and successively change this by \textbf{B}: choosing $k = 5$ with constant weights (unweighted); \textbf{C}: setting $k$ and the approximation type (for FRNN) through leave-one-out validation; \textbf{D}: choosing Boscovich distance; \textbf{E}: choosing Samworth distance- and rank-weights; and \textbf{F}: choosing $r_2$-scaling. Dashed lines: median; dotted lines: first and third quartiles.}
\label{fig_violins}
\end{figure}

\begin{table*}
\centering
\caption{Mean 5-fold cross-validation AUROC with Boscovich distance, $r_2$-scaling and Samworth distance-weights (FNN) or both Samworth distance- and rank-weights (NN, FRNN). \textbf{{Bold}}: highest value (before rounding).}
\label{tab_auroc}
\begin{tabular}{llllllll}
\toprule
dataset & NN & FNN & FRNN & dataset & NN & FNN & FRNN \\
\midrule
accent & \bftab 0.976 & 0.973 & 0.971 & mfeat & 0.999 & 0.999 & \bftab 1.000 \\
acoustic-features & 0.930 & 0.921 & \bftab 0.938 & miniboone & 0.962 & 0.954 & \bftab 0.962 \\
ai4i2020 & \bftab 0.957 & 0.911 & 0.954 & new-thyroid & 0.989 & 0.985 & \bftab 0.994 \\
alcohol & \bftab 1.000 & \bftab 1.000 & \bftab 1.000 & oral-toxicity & \bftab 0.888 & 0.871 & 0.878 \\
androgen-receptor & 0.855 & 0.847 & \bftab 0.867 & page-blocks & 0.975 & 0.966 & \bftab 0.982 \\
avila & 0.986 & 0.979 & \bftab 0.987 & phishing-websites & \bftab 0.996 & 0.994 & 0.994 \\
banknote & 1.000 & 1.000 & \bftab 1.000 & plrx & 0.479 & 0.442 & \bftab 0.496 \\
bioaccumulation & 0.741 & 0.733 & \bftab 0.755 & pop-failures & 0.947 & \bftab 0.949 & 0.941 \\
biodeg & \bftab 0.926 & 0.903 & 0.926 & post-operative & 0.476 & 0.478 & \bftab 0.479 \\
breasttissue & \bftab 0.932 & 0.902 & 0.921 & qualitative-bankruptcy & \bftab 1.000 & 0.995 & \bftab 1.000 \\
ca-cervix & \bftab 0.949 & 0.910 & \bftab 0.949 & raisin & \bftab 0.927 & 0.909 & 0.925 \\
caesarian & \bftab 0.708 & 0.708 & 0.685 & rejafada & \bftab 0.958 & 0.950 & 0.957 \\
ceramic & 0.758 & 0.765 & \bftab 0.800 & rice & 0.979 & 0.964 & \bftab 0.979 \\
cmc & 0.701 & 0.627 & \bftab 0.701 & seeds & 0.991 & 0.979 & \bftab 0.991 \\
codon-usage & 0.969 & 0.967 & \bftab 0.975 & segment & \bftab 0.998 & 0.997 & 0.998 \\
coimbra & 0.774 & 0.760 & \bftab 0.794 & seismic-bumps & 0.767 & 0.673 & \bftab 0.773 \\
column & \bftab 0.919 & 0.905 & 0.915 & sensorless & 1.000 & 0.999 & \bftab 1.000 \\
debrecen & \bftab 0.734 & 0.676 & 0.734 & sepsis-survival & 0.700 & \bftab 0.701 & 0.666 \\
dermatology & \bftab 0.999 & 0.998 & 0.999 & shuttle & 0.998 & 0.997 & \bftab 1.000 \\
diabetes-risk & 0.998 & 0.991 & \bftab 0.999 & skin & 1.000 & 1.000 & \bftab 1.000 \\
divorce & \bftab 1.000 & \bftab 1.000 & \bftab 1.000 & somerville & 0.595 & 0.561 & \bftab 0.614 \\
dry-bean & \bftab 0.996 & 0.994 & 0.994 & sonar & 0.959 & 0.927 & \bftab 0.961 \\
ecoli & 0.969 & 0.947 & \bftab 0.972 & south-german-credit & 0.789 & 0.756 & \bftab 0.794 \\
electrical-grid & \bftab 0.968 & 0.959 & 0.968 & spambase & \bftab 0.980 & 0.977 & 0.977 \\
faults & 0.959 & 0.950 & \bftab 0.962 & spectf & 0.841 & 0.837 & \bftab 0.852 \\
fertility & 0.687 & \bftab 0.763 & 0.647 & sportsarticles & 0.880 & 0.861 & \bftab 0.884 \\
flowmeters & \bftab 0.977 & 0.961 & 0.970 & sta-dyn-lab & 0.994 & 0.992 & \bftab 0.996 \\
forest-types & 0.969 & 0.958 & \bftab 0.974 & tcga-pancan-hiseq & 1.000 & 1.000 & \bftab 1.000 \\
gender-gap & \bftab 0.704 & 0.623 & 0.700 & thoraric-surgery & 0.624 & 0.545 & \bftab 0.630 \\
glass & 0.942 & 0.917 & \bftab 0.948 & transfusion & 0.725 & 0.689 & \bftab 0.747 \\
haberman & 0.681 & 0.633 & \bftab 0.709 & tuandromd & 0.997 & 0.998 & \bftab 0.999 \\
hcv & 0.993 & 0.992 & \bftab 0.994 & urban-land-cover & 0.973 & 0.968 & \bftab 0.974 \\
heart-failure & \bftab 0.867 & 0.863 & 0.863 & vehicle & 0.912 & 0.887 & \bftab 0.913 \\
house-votes-84 & \bftab 0.984 & 0.979 & 0.983 & warts & \bftab 0.907 & 0.870 & 0.899 \\
htru2 & 0.972 & 0.963 & \bftab 0.977 & waveform & \bftab 0.972 & 0.971 & 0.971 \\
ilpd & 0.727 & 0.675 & \bftab 0.734 & wdbc & 0.989 & 0.989 & \bftab 0.994 \\
ionosphere & 0.942 & 0.935 & \bftab 0.982 & wifi & \bftab 1.000 & 0.999 & 0.999 \\
iris & 0.997 & \bftab 0.998 & 0.997 & wilt & 0.962 & 0.940 & \bftab 0.974 \\
landsat & \bftab 0.990 & 0.987 & 0.989 & wine & \bftab 1.000 & \bftab 1.000 & \bftab 1.000 \\
leaf & \bftab 0.975 & 0.961 & 0.975 & wisconsin & 0.990 & 0.991 & \bftab 0.996 \\
letter & 0.999 & 0.999 & \bftab 0.999 & wpbc & 0.589 & 0.573 & \bftab 0.665 \\
lrs & \bftab 0.915 & 0.885 & 0.899 & yeast & 0.869 & 0.860 & \bftab 0.890 \\
magic & 0.916 & 0.893 & \bftab 0.923 &  &  &  &  \\
\bottomrule
\end{tabular}
\end{table*}

In Subsection~\ref{sec_results_fnn}, we observed that FNN performs best on our data with Samworth distance-weights, but that the difference with respect to reciprocally linear and reciprocally square distance-weights is not significant for all scaling-types. However, when we compare FNN to NN, we find that for all three kernels, NN performs significantly better (Table~\ref{tab_p_values_nn_vs_fnn}).

For both NN and FRNN, we obtained the best results with a combination of Samworth distance- and rank-weights. When we compare NN and FRNN against each other, we find that FRNN performs better (Table~\ref{tab_p_values_frnn_vs_nn}).

Table~\ref{tab_auroc} lists the mean AUROC achieved with NN, FNN and FRNN for each dataset with $r_2$-scaling (the results are comparable for the other scaling types). Fig.~\ref{fig_violins} illustrates the increase in performance for NN and FRNN due to the various hyperparameter choices proposed in this paper.

\subsection{Distance measures}
\label{sec_results_distance_measures}

At the start of Subsections~\ref{sec_results_nn} and \ref{sec_results_frnn}, we found that, regardless of weights and scaling, NN and FRNN generally perform better with Boscovich distance than with Euclidean and Chebyshev distance. We now consider whether, despite this general trend, there might be types of datasets for which Euclidean and Chebyshev distance present an advantage.\footnote{Note that we obtain similar results regarding the performance of different weight types and scaling measures for Euclidean and Chebyshev distance as we did for Boscovich distance.}

\begin{figure}
\centering
\includegraphics[width=\linewidth]{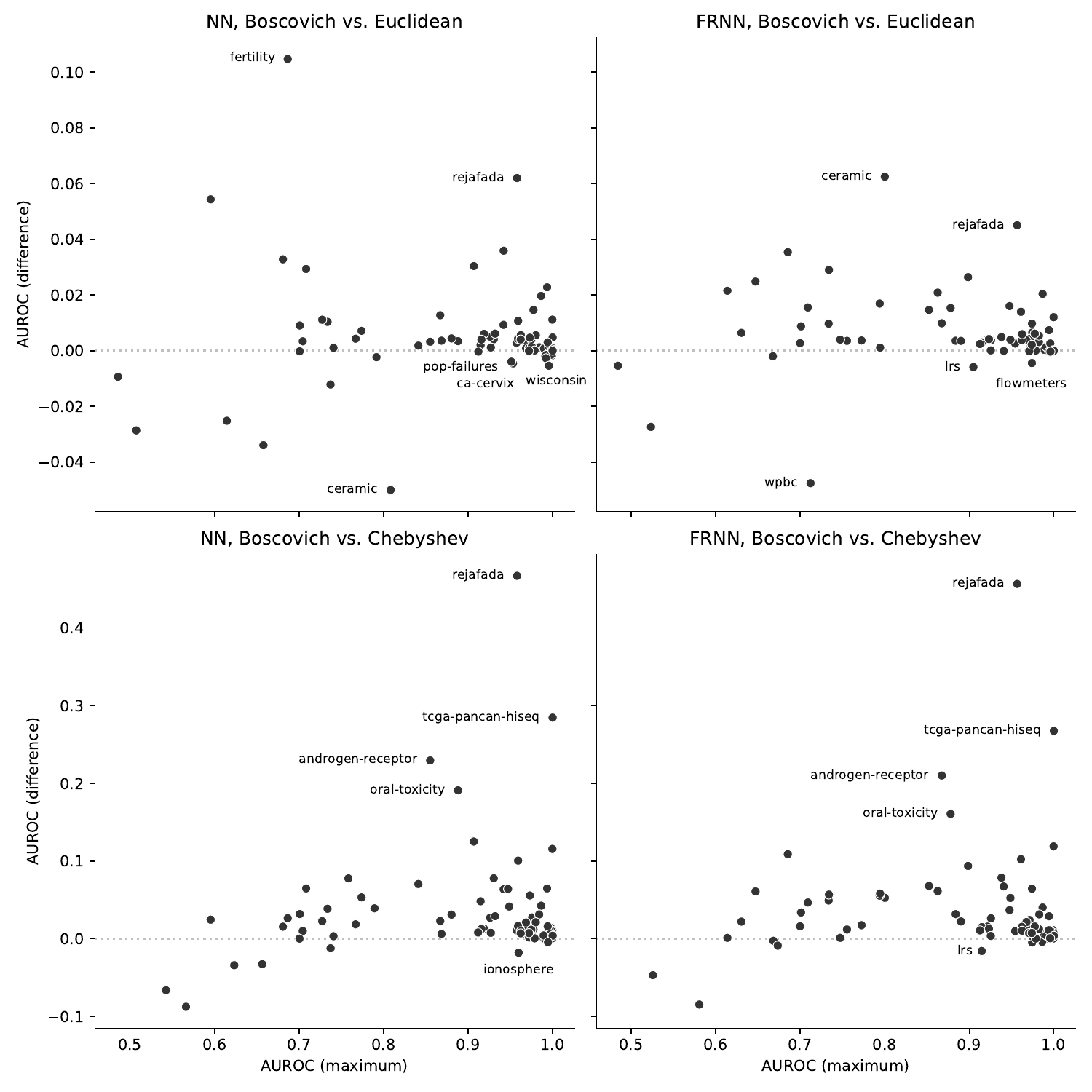}
\caption{AUROC difference (y-axis) between Boscovich distance and Euclidean (top row) or Chebyshev (bottom row) distance, offset against the maximum of the two AUROC scores (x-axis), for NN (left column) and FRNN (right column), with Samworth distance- and rank-weights and $r_2$-scaling, applied to the datasets in this paper.}
\label{fig_metric_differences}
\end{figure}

In Fig.\ref{fig_metric_differences}, we compare the performance of NN and FRNN with Boscovich distance against Euclidean and Chebyshev distance. For datasets above the baseline, classification performance is higher for Boscovich distance, whereas for datasets below the baseline, performance is higher for Euclidean or Chebyshev distance. The x-axis expresses how difficult the classification problem is.

The datasets below the baseline, for which Euclidean or Chebyshev distance outperforms Boscovich distance, mostly fall into two subgroups. The datasets on the left side of the plot are very difficult to classify, which introduces a degree of randomness, making even substantial performance differences between distance measures not very meaningful. The datasets on the right side of the plot below the baseline are still very close to it, meaning that the performance difference is very small. Overall, we are not able to identify a specific dataset type for which Euclidean or Chebyshev distance possesses a specific advantage.

On the other hand, we note that Chebyshev distance in particular does lead to substantially poorer performance on the \e{rejafada}, \e{tcga-pancan-hiseq}, \e{androgen-receptor} and \e{oral-toxicity} datasets. These are the four datasets in our collection with the highest dimensionality. Because the Chebyshev distance between two records is completely determined by the attribute with the highest difference, it effectively discards a large part of the information contained in these datasets.

\subsection{Yager-$\frac{1}{2}$ weights}

We now consider the results of the Yager-$\frac{1}{2}$ kernel that we have proposed. When we equip NN with both Yager-$\frac{1}{2}$ distance- and rank-weights, this performs slightly better on our data than Samworth distance- and rank-weights, but the difference is not significant ($p < 0.25$ across scaling measures). Interestingly, unlike the Samworth kernel, the Yager-$\frac{1}{2}$ kernel appears to perform about as well when only used for distance-weights and when used for both distance- and rank-weights. Correspondingly, Yager-$\frac{1}{2}$ distance-weights perform significantly better than Samworth distance-weights ($p < 0.0053$). Thus the main advantage of the Yager-$\frac{1}{2}$ kernel is that it enables comparable performance as the Samworth kernel, but is easier to implement, because it does not require the addition of rank-weights and because it is not dependent on the dimensionality of the dataset.

In contrast, for FNN we obtain comparable performance between Samworth and Yager-$\frac{1}{2}$ distance-weights with $r_1$- or $r_2$-scaling, and for FRNN we find that Samworth distance- and rank-weights still perform significantly better than Yager-$\frac{1}{2}$ distance- and rank-weights ($p < 0.050$ across scaling measures).

\subsection{Weight optimisation}

Finally, we ask whether it is beneficial to optimise not just the number of neighbours $k$ through leave-one-out validation, but also the weight type. Because we obtained our best results with Samworth and Yager weights, we focus on these two weight types for this experiment. In addition, we restrict ourselves to Boscovich distance and $r_2$-scaling.

Recall that the Yager kernel is parametrised by $p$, which controls the relative steepness of the function for values close to 0 and close to 1 (Fig.~\ref{fig_kernels}). Thus, we can optimise the weight type by optimising $p$. Similarly, the Samworth kernel is parametrised by the value $m$. Until now, we have used the dimensionality of the data to set $m$, but we can also treat it as an independent, optimisable, parameter.

In order to aid the parameter search, we reparametrise the Yager and Samworth kernels by substituting $p = 2^q$ and $m = 2^{-(q + 1)}$, resulting, respectively, in $a \rma (1 - a^{2^q})^{2^{-q}}$ and $a \rma 1 - a^{2^q}$. Because we now optimise two hyperparameters, and because $q$ is continuous, unlike $k$, we can no longer perform an exhaustive search. Therefore, we perform Malherbe-Powell optimisation \cite{malherbe17global,powell09bobyqa,king17global} as implemented in the \e{dlib} library \cite{king09dlibml}, which was the best-performing algorithm for a series of similar hyperparameter optimisation problems \cite{lenz22optimised}, with a search budget of 100 evaluations.

We find that the effectiveness of optimising $q$ depends on a proper delineation of the search space. We originally considered values in $[-10, 10]$, however, for the Yager weights most optimised values end up in the relatively narrow interval $[-4, 2]$, while for the Samworth weights, there is a large concentration of values at -10, suggesting that their true optimal value is lower. With this search space, optimised Samworth and Yager weights do not outperform their unoptimised equivalents for NN.

However, when we change the search space to $[-4, 2]$ for Yager weights and $[-20, 0]$ for Samworth weights, we obtain better results. Now, both for a combination of Samworth distance and rank weights ($p = 0.045$) as well as for Yager distance weights ($p = 0.058$), optimisation increases performance for NN. For FRNN, optimising Samworth distance- and rank weights also increases performance, but the difference is not significant ($p = 0.22$). However, FRNN still outperforms NN with optimised Samworth distance and rank weights ($p = 0.038$).

\section{Conclusion}
\label{sec_nn_conclusion}

In this paper, we have provided a comprehensive overview of the different weighting variants of NN, FNN and FRNN classification that have been proposed in the literature. We have proposed a uniform framework for these proposals and conducted an evaluation on 85 real-life datasets. This allows us to draw the following conclusions:

\begin{itemize}
 \item Weighting can be expressed as the application of a kernel function to the distances and/or ranks of the nearest neighbours of a test record --- we have provided an overview of kernel functions that correspond to existing weighting proposals in the literature.
 \item In particular, Samworth rank-weights, which have been shown to be theoretically optimal, converge to a kernel function that depends on the dimensionality of the data, and that can also be applied to obtain distance-weights.
 \item On real-life datasets, both NN and FRNN perform better with a combination of Samworth rank- and distance-weights than with other weight types proposed in the literature, while FNN appears to perform best with Samworth distance-weights and constant rank-weights.
 \item However, NN and FNN appear to perform equally well with Yager-$\frac{1}{2}$ weights, a novel weight type inspired by fuzzy Yager negation. For NN, the Yager-$\frac{1}{2}$ kernel offers two practical benefits over the Samworth kernel: it only needs to be applied to obtain distance-weights (not rank-weights), and it does not depend on the dimensionality of the dataset. For FRNN, Samworth weights still perform better.
 \item For NN in particular, performance can be further increased by optimising not just $k$, but also the weight type, for which we can use the parameter $p$ of the Yager kernel and the parameter $m$ of the Samworth kernel.
 \item Boscovich distance clearly outperforms Euclidean and Chebyshev distance, regardless of other hyperparameter choices. Chebyshev distance is a particularly bad choice for very high-dimensional datasets.
 \item With Samworth and Yager-$\frac{1}{2}$ weights, rescaling attributes by $r_1$ (mean absolute deviation around the median), $r_2$ (standard deviation) or $r_{\infty}^*$ (semi-interquartile range) produces comparable results, while these are all better than rescaling by $r_{\infty}$ (half-range).
 \item Our comparison between NN and FNN with identical distance-weights reveals that in practice, the fuzzification of class membership degrees in FNN leads to systematically lower performance. In contrast, with its more fundamentally different approach, FRNN does generally outperform NN when both are equipped with their best-performing weighting scheme (Samworth distance- and rank-weights).
\end{itemize}

We believe that these results serve as a useful baseline for future applications and research. For applications, we recommend the use of FRNN classification with Samworth rank- and distance-weights, Boscovich distance, and any one of $r_1$-, $r_2$- or $r_{\infty}^*$-scaling, while $k$ can be optimised through efficient leave-one-out validation. With the classical NN algorithm, we recommend the same hyperparameter choices, except that Yager-$\frac{1}{2}$ distance-weights may be substituted and rank-weights omitted.

We suggest that future research should concentrate on identifying even better-performing kernel functions. For this, the contours of the Samworth and Yager-$\frac{1}{2}$ kernels may serve as a useful starting point. In particular, we hope that in this way the following two questions may be answered:

\begin{itemize}
 \item Should an optimal kernel depend on the dimensionality of the data? The good performance of the Samworth kernel suggests that the answer is yes. However, we note that its profile is actually very similar for typical dimensionalities of $4 \leq m \leq 256$, so that its variation according to dimensionality may be far less important than the basic outline of its profile. This is seemingly confirmed by the relatively strong performance of the Yager-$\frac{1}{2}$ kernel, which has a similar profile.
 \item For NN, do we need both distance- and rank-weights? The fact that Yager-$\frac{1}{2}$ distance-weights alone perform as well as a combination of Samworth distance- and rank-weights suggests that with the right kernel, we can do away with rank-weights.
\end{itemize}

We hope that any future evaluations of competing proposals on tabular data will be based on a similar collection of real-life classification problems to the one that we have assembled, which we have made available for this purpose. Finally, it would also be interesting to investigate whether the same weighting kernels perform well for nearest neighbour classification of non-tabular data, like images and texts.

\section*{CRediT authorship contribution statement}

\textbf{Oliver Urs Lenz}: Conceptualisation, Software, Formal analysis, Investigation, Data Curation, Writing --- Original Draft. \textbf{Henri Bollaert}: Conceptualisation, Investigation, Writing --- Review \& editing. \textbf{Chris Cornelis}: Writing --- Review \& editing, Supervision.

\section*{Declaration of competing interest}

The authors declare that they have no known competing financial interests or personal relationships that could have appeared to influence the work reported in this paper.

\section*{Data availability}
The datasets used in this paper can be downloaded from \url{https://cwi.ugent.be/~oulenz/datasets/lenz-2024-unified.tar.gz}.

\section*{Acknowledgments}
\label{sec_acknowledgement}
The research reported in this paper was conducted with the financial support of the Odysseus programme of the Research Foundation -- Flanders (FWO).

This publication is part of the project Digital Twin with project number P18-03 of the research programme TTW Perspective, which is (partly) financed by the Dutch Research Council (NWO).

We thank Martine De Cock for raising the important question whether NN and FRNN weighting could be unified, which sparked our inspiration for the present framework.

\appendix

\section{Proofs that existing weight types can be represented through kernels}
\label{sec_proofs}

In this appendix we will show how most existing weighting proposals discussed in Subsection~\ref{sec_weighted_nearest_neighbour_classification} can be rewritten as kernel functions applied to $i^* = \frac{i}{k+1}$ (rank-weighting) or $d_i^* = \frac{d_i}{d_k}$ (distance-weighting), making them compatible with our generalised definition of NN classification \eqref{eq_nn_new}. Firstly, note that both in the original equation for NN \eqref{eq_wnn}, and in our generalised equation \eqref{eq_nn_new}, we rescale each class score by the total sum of the weights. Therefore, we do not require that weights sum to 1. Moreover, multiplying all weights by a positive constant produces identical class scores. We will use the proportionality symbol $\propto$ to indicate that two sets of weights only differ by a positive constant factor.

\subsection{Linear weights}
The linear distance-weights proposed by Dudani \cite{dudani73experimental} can be simplified as follows:

 \begin{equation*}
\begin{aligned}
 s_i &= \frac{d_k - d_i}{d_k - d_1}\\
 &\propto \frac{d_k - d_i}{d_k - d_1} \cdot \frac{d_k - d_1}{d_k}\\
 &= 1 - d_i^*.
 \end{aligned}
 \end{equation*}
Similarly, for the linear rank-weights proposed by Dudani \cite{dudani73experimental}, we have:

 \begin{equation*}
\begin{aligned}
 w_i &= k - i + 1\\
 &\propto \frac{k - i + 1}{k + 1}\\
 &= 1 - i^*.
 \end{aligned}
 \end{equation*}
And likewise for the linear rank-weights proposed by Stone \cite{stone77consistent}:

 \begin{equation*}
\begin{aligned}
 w_i &= \frac{k - i + 1}{k(k+1)/2}\\
 &\propto \frac{k - i + 1}{k + 1}\\
 &= 1 - i^*.
 \end{aligned}
 \end{equation*}

\subsection{Sugeno weights}
In a similar way, we can simplify the weights proposed by Gou et al. \cite{gou12new}:

 \begin{equation}
 \label{eq_gou_sugeno}
\begin{aligned}
 s_i &= \frac{d_k - d_i}{d_k - d_1} \cdot \frac{d_k + d_1}{d_k + d_i}\\
 &\propto \frac{d_k - d_i}{d_k - d_1} \cdot \frac{d_k + d_1}{d_k + d_i} \cdot \frac{d_k - d_1}{d_k + d_1} \cdot \frac{1}{d_k}\\
 &= \frac{1 - d_i^*}{1 + d_i^*}.
 \end{aligned}
 \end{equation}
We call this the \e{Sugeno-1 kernel}, because it is in fact an instance of fuzzy Sugeno negation \cite{sugeno73constructing}, with $\lambda = 1$:

 \begin{equation*}
 \label{eq_sugeno}
a \lrma \frac{1 - a}{1 + \lambda a}
 \end{equation*}

\subsection{Quadratic weights}
For the quadratic rank-weights proposed by Stone \cite{stone77consistent}, we are left with a remainder, but this vanishes as $k$ increases:

 \begin{equation*}
\begin{aligned}
 w_i &= \frac{k^2 - (i - 1)^2}{k(k+1)(4k-1)/6}\\
 &\propto \frac{k^2 - (i - 1)^2}{(k + 1)^2}\\
 &= 1 - \frac{2k + i^2 - 2i + 2}{(k + 1)^2}\\
 &= 1 - (i^*)^2 - \frac{2(k - i + 1)}{(k + 1)^2}\\
 &\ra 1 - (i^*)^2 \text{ as } k \ra \infty.
 \end{aligned}
 \end{equation*}

\subsection{Reciprocally linear weights}
The reciprocally linear distance-weights proposed by Dudani \cite{dudani73experimental} are proportional to a reciprocally linear kernel:

 \begin{equation*}
 \begin{aligned}
 \label{eq_reci_kernel}\\
 s_i &= \frac{1}{d_i}\\
 &\propto \frac{d_k}{d_i}\\
 &= \frac{1}{d_i^*}.
 \end{aligned}
 \end{equation*}
Note that this kernel is improper, as its value goes to $\infty$ as $d_i^*$ goes to 0.

Likewise, for the reciprocally linear rank-weights proposed by Gou et al. \cite{gou11novel}, we have:

\begin{equation*}
\begin{aligned}
 w_i &= \frac{1}{i}\\
 &\propto \frac{k+1}{i}\\
 &= \frac{1}{i^*}.
 \end{aligned}
\end{equation*}

\subsection{Reciprocally quadratic weights}
Analogously, the reciprocally quadratic distance-weights proposed by Shepard \cite{shepard68twodimensional} are proportional to a reciprocally quadratic kernel, which is also improper:

 \begin{equation*}
 \begin{aligned}
 \label{eq_reci_2_kernel}
 s_i &= \frac{1}{d_i^2}\\
 &\propto \frac{d_k^2}{d_i^2}\\
 &= \frac{1}{\left(d_i^*\right)^2}.
 \end{aligned}
 \end{equation*}

\subsection{Macleod weights}
We can also simplify the distance-weights proposed by Macleod et al. \cite{macleod87reexamination}:

 \begin{equation*}
\begin{aligned}
 s_i &= \frac{d_k - d_i + d_k - d_1}{2(d_k - d_1)}\\
 &\propto \frac{d_k - d_i + d_k - d_1}{2(d_k - d_1)} \cdot \frac{2(d_k - d_1)}{d_k}\\
 &= 2 - d_i^* - d_1^*.
 \end{aligned}
 \end{equation*}
However, the resulting function still depends on $d_1$. That means that we can use this function to calculate distance-weights, but it does not generalise to a kernel that we could also apply to rank-weights.

\subsection{Laplace weights}
The only remaining distance-weights that cannot be rewritten into a kernel function are the Laplace weights $e^{-d_i}$ proposed in \cite{zavrel97empirical}. Note that these are not homogeneous, i.e. the weighting depends on the absolute scale of the distances, which is arguably undesirable. However, we can consider the Laplace kernel $e^{-d_i^*}$.

\subsection{Samworth weights}
For the Samworth weights, if we fix a particular value $k > 0$, and write $h = \frac{1}{k+1}$ (for reasons of space), we can define the following kernel function $f^k$ such that $f^k(\frac{i}{k + 1})$ is the $i$th weight as in \eqref{eq_samworth}:

\begin{equation*}
 f^k(a) = \frac{1}{k} \left(1 + \frac{m}{2} - \frac{m}{2k^{\frac{2}{m}}}\left((a/h)^{1 + \frac{2}{m}} - (a/h - 1)^{1 + \frac{2}{m}}\right)\right).
\end{equation*}
$f^k$ is proportional to $\frac{2k}{m + 2} \cdot f^k$, and we have the following lemma:

\begin{lemma}

 \begin{equation*}
 \lim_{k \ra \infty} \frac{2k}{m + 2} \cdot f^k = 1 - a^{\frac{2}{m}}.
 \end{equation*}

\end{lemma}

\begin{proof}
{\allowdisplaybreaks
\begin{align*}
 &\mathrel{\phantom{=}} \lim_{k \ra \infty} \frac{2k}{m + 2} \cdot f_k\\
 &= \lim_{k \ra \infty} 1 - \frac{m}{m + 2} \cdot \frac{1}{k^{\frac{2}{m}}}\left((a/h)^{1 + \frac{2}{m}} - (a/h - 1)^{1 + \frac{2}{m}}\right)\\
 &= 1 - \frac{m}{m + 2} \cdot \lim_{k \ra \infty} \frac{(a/h)^{1 + \frac{2}{m}} - (a/h - 1)^{1 + \frac{2}{m}}}{k^{\frac{2}{m}}}\\
 &= 1 - \frac{m}{m + 2} \cdot \lim_{k \ra \infty} \frac{h^{1 + \frac{2}{m}}}{h^{1 + \frac{2}{m}}}\frac{(a/h)^{1 + \frac{2}{m}} - (a/h - 1)^{1 + \frac{2}{m}}}{k^{\frac{2}{m}}}\\
 &= 1 - \frac{m}{m + 2} \cdot \lim_{k \ra \infty} \frac{a^{1 + \frac{2}{m}} - (a - h)^{1 + \frac{2}{m}}}{h \cdot \frac{k^{2/m}}{(k+1)^{2/m}}}\\
 &= 1 - \frac{m}{m + 2} \cdot \lim_{h \ra 0} \frac{a^{1 + \frac{2}{m}} - (a - h)^{1 + \frac{2}{m}}}{h}\\
 &= 1 - \frac{m}{m + 2} \cdot \left(1 + \frac{2}{m}\right)a^{\frac{2}{m}}\tag{*}\\
 &= 1 - a^{\frac{2}{m}},
 \end{align*}
}
where $(*)$ is the polynomial rule for derivation.
\end{proof}
Accordingly, we call $1 - a^{\frac{2}{m}}$ the Samworth kernel. Note that the Samworth kernel is a fuzzy negation, since it sends $a = 1$ to 0.

\subsection{Fibonacci weights}
The only rank-weights that cannot be obtained by applying a kernel function are the Fibonacci weights from Pao et al. \cite{pao07comparative}, because their relative distribution depends on $k$.

\bibliographystyle{elsarticle-num}
\bibliography{20231127_nn_variants}

\begin{thebibliography}{10}
\expandafter\ifx\csname url\endcsname\relax
  \def\url#1{\texttt{#1}}\fi
\expandafter\ifx\csname urlprefix\endcsname\relax\def\urlprefix{URL }\fi
\expandafter\ifx\csname href\endcsname\relax
  \def\href#1#2{#2} \def\path#1{#1}\fi

\bibitem{fix51discriminatory}
E.~Fix, J.~L. Hodges, Jr, Discriminatory analysis --- nonparametric
  discrimination: Consistency properties, Technical report 21-49-004, {USAF}
  School of Aviation Medicine, Randolph Field, Texas (1951).

\bibitem{cunningham21knearest}
P.~Cunningham, S.~J. Delany, {k-Nearest} neighbour classifiers --- {A}
  tutorial, ACM Comput. Surv. 54~(6), art. no. 128 (2021).

\bibitem{james23statistical}
G.~James, D.~Witten, T.~Hastie, R.~Tibshirani, J.~Taylor, An Introduction to
  Statistical Learning: with Applications in Python, Springer, Cham, 2023, pp.
  36--39, 164--166.

\bibitem{arian20protein}
R.~Arian, A.~Hariri, A.~Mehridehnavi, A.~Fassihi, F.~Ghasemi, Protein kinase
  inhibitors' classification using {K-Nearest} neighbor algorithm, Comput.
  Biol. Chemistry 86, art. no. 107269 (2020).

\bibitem{kour22visionbased}
N.~Kour, S.~Gupta, S.~Arora, A vision-based clinical analysis for
  classification of knee osteoarthritis, {P}arkinson's disease and normal gait
  with severity based on k-nearest neighbour, Expert Syst. 39~(6), art. no.
  e12955 (2022).

\bibitem{shahrestani23developing}
S.~Shahrestani, A.~K. Chan, E.~F. Bisson, M.~Bydon, S.~D. Glassman, K.~T.
  Foley, C.~I. Shaffrey, E.~A. Potts, M.~E. Shaffrey, D.~Coric, J.~J. Knightly,
  P.~Park, M.~Y. Wang, K.-M. Fu, J.~R. Slotkin, A.~L. Asher, M.~S. Virk, G.~D.
  Michalopoulos, J.~Guan, R.~W. Haid, N.~Agarwal, D.~Chou, P.~V. Mummaneni,
  Developing nonlinear k-nearest neighbors classification algorithms to
  identify patients at high risk of increased length of hospital stay following
  spine surgery, Neurosurgical Focus 54~(6), art. no. E7 (2023).

\bibitem{cosenza21comparison}
D.~N. Cosenza, L.~Korhonen, M.~Maltamo, P.~Packalen, J.~L. Strunk,
  E.~N{\ae}sset, T.~Gobakken, P.~Soares, M.~Tom{\'e}, Comparison of linear
  regression, k-nearest neighbour and random forest methods in airborne
  laser-scanning-based prediction of growing stock, Forestry 94~(2) (2021)
  311--323.

\bibitem{gomezgil24vibrationbased}
F.~J. Gomez-Gil, V.~Mart{\'\i}nez-Mart{\'\i}nez, R.~Ruiz-Gonzalez,
  L.~Mart{\'\i}nez-Mart{\'\i}nez, J.~Gomez-Gil, Vibration-based monitoring of
  agro-industrial machinery using a k-{N}earest {N}eighbors ({kNN}) classifier
  with a {H}armony {S}earch ({HS}) frequency selector algorithm, Computers and
  Electronics in Agriculture 217, art. no. 108556 (2024).

\bibitem{martinmartin23using}
M.~Mart{\'\i}n-Mart{\'\i}n, M.~Bullejos, D.~Cabezas, F.~J. Alcal{\'a}, Using
  python libraries and k-{N}earest neighbors algorithms to delineate
  syn-sedimentary faults in sedimentary porous media, Marine and Petroleum
  Geology 153, art. no. 106283 (2023).

\bibitem{suleymanov23spatial}
A.~Suleymanov, I.~Tuktarova, L.~Belan, R.~Suleymanov, I.~Gabbasova,
  L.~Araslanova, Spatial prediction of soil properties using random forest,
  k-nearest neighbors and cubist approaches in the foothills of the {U}ral
  {M}ountains, {R}ussia, Modeling Earth Systems and Environment 9~(3) (2023)
  3461--3471.

\bibitem{aslinezhad20turbine}
M.~Aslinezhad, M.~A. Hejazi, Turbine blade tip clearance determination using
  microwave measurement and k-nearest neighbour classifier, Measurement 151,
  art. no. 107142 (2020).

\bibitem{konieczny21use}
J.~Konieczny, J.~Stojek, Use of the $k$-nearest neighbour classifier in wear
  condition classification of a positive displacement pump, Sensors 21~(18),
  art. no. 6247 (2021).

\bibitem{shijer24evaluating}
S.~S. Shijer, A.~H. Jassim, L.~A. Al-Haddad, T.~T. Abbas, Evaluating electrical
  power yield of photovoltaic solar cells with k-{N}earest neighbors: A machine
  learning statistical analysis approach, e-Prime --- Advances in Electrical
  Engineering, Electronics and Energy 9, art. no. 100674 (2024).

\bibitem{khandelwal21nearest}
U.~Khandelwal, A.~Fan, D.~Jurafsky, L.~Zettlemoyer, M.~Lewis, Nearest neighbor
  machine translation, in: Proc. 9th Int. Conf. Learn. Representations, 2021.

\bibitem{kaminska23fuzzyanalysis}
O.~Kaminska, C.~Cornelis, V.~Hoste, Fuzzy rough nearest neighbour methods for
  aspect-based sentiment analysis, Electronics 12~(5), art. no. 1088 (2023).

\bibitem{kaminska23fuzzyirony}
O.~Kaminska, C.~Cornelis, V.~Hoste, Fuzzy rough nearest neighbour methods for
  detecting emotions, hate speech and irony, Inf. Sci. 625 (2023) 521--535.

\bibitem{maillo20fast}
J.~Maillo, S.~Garc{\'\i}a, J.~Luengo, F.~Herrera, I.~Triguero, Fast and
  scalable approaches to accelerate the fuzzy k-nearest neighbors classifier
  for big data, {IEEE} Trans. Fuzzy Syst. 28~(5) (2020) 874--886.

\bibitem{lenz20scalable}
O.~U. Lenz, D.~Peralta, C.~Cornelis, Scalable approximate {FRNN-OWA}
  classification, {IEEE} Trans. Fuzzy Syst. 28~(5) (2020) 929--938.

\bibitem{shokrzade21novel}
A.~Shokrzade, M.~Ramezani, F.~A. Tab, M.~A. Mohammad, A novel extreme learning
  machine based {kNN} classification method for dealing with big data, Expert
  Syst. Appl. 183, art. no. 115293 (2021).

\bibitem{keller85fuzzy}
J.~M. Keller, M.~R. Gray, J.~A. Givens, A fuzzy $k$-nearest neighbor algorithm,
  IEEE Trans. Syst., Man, Cybern.~(4) (1985) 580--585.

\bibitem{jensen08new}
R.~Jensen, C.~Cornelis, A new approach to fuzzy-rough nearest neighbour
  classification, in: Proc. 6th Int. Conf. Rough Sets Current Trends Comput.,
  2008, pp. 310--319.

\bibitem{samworth12optimal}
R.~J. Samworth, Optimal weighted nearest neighbour classifiers, Ann. Statist.
  (2012) 2733--2763.

\bibitem{gou12new}
J.~Gou, L.~Du, Y.~Zhang, T.~Xiong, et~al., A new distance-weighted k-nearest
  neighbor classifier, J. Inf. Comput. Sci. 9~(6) (2012) 1429--1436.

\bibitem{watson64smooth}
G.~S. Watson, Smooth regression analysis, Sankhy{\=a}: Indian J. Statist., Ser.
  A (1964) 359--372.

\bibitem{royall66class}
R.~M. Royall, A class of non-parametric estimates of a smooth regression
  function., Ph.D. thesis (1966).

\bibitem{shepard68twodimensional}
D.~Shepard, A two-dimensional interpolation function for irregularly-spaced
  data, in: Proc. 1968 23rd ACM Nat. Conf., 1968, pp. 517--524.

\bibitem{rosenblatt56remarks}
M.~Rosenblatt, Remarks on some nonparametric estimates of a density function,
  Ann. Math. Statist. (1956) 832--837.

\bibitem{dudani73experimental}
S.~A. Dudani, An experimental study of moment methods for automatic
  identification of three-dimensional objects from television images, Ph.D.
  thesis, The Ohio State University (1973).

\bibitem{dudani76distance}
S.~A. Dudani, The distance-weighted $k$-nearest-neighbor rule, IEEE Trans.
  Syst., Man, Cybern. 6~(4) (1976) 325--327.

\bibitem{stone77consistent}
C.~J. Stone, Consistent nonparametric regression, Ann. Statist. (1977)
  595--620.

\bibitem{altman92introduction}
N.~S. Altman, An introduction to kernel and nearest-neighbor nonparametric
  regression, Amer. Statistician 46~(3) (1992) 175--185.

\bibitem{gou11novel}
J.~Gou, T.~Xiong, Y.~Kuang, A novel weighted voting for k-nearest neighbor
  rule., J. Comput. 6~(5) (2011) 833--840.

\bibitem{pao07comparative}
T.-L. Pao, Y.-T. Chen, J.-H. Yeh, Y.-M. Cheng, Y.-Y. Lin, A comparative study
  of different weighting schemes on {KNN}-based emotion recognition in
  {M}andarin speech, in: 3rd Int. Conf. Intell. Comput., 2007, pp. 997--1005.

\bibitem{shepard87toward}
R.~N. Shepard, Toward a universal law of generalization for psychological
  science, Science 237~(4820) (1987) 1317--1323.

\bibitem{zavrel97empirical}
J.~Zavrel, An empirical re-examination of weighted voting for {$k$-NN}, in:
  Proc. 7th Belg.-Dutch Conf. Mach. Learn., 1997, pp. 139--145.

\bibitem{bailey78note}
T.~Bailey, A.~Jain, A note on distance-weighted $k$-nearest neighbor rules,
  IEEE Trans. Syst., Man, Cybern. 8~(4) (1978) 311--313.

\bibitem{macleod87reexamination}
J.~E. Macleod, A.~Luk, D.~M. Titterington, A re-examination of the
  distance-weighted k-nearest neighbor classification rule, IEEE Trans. Syst.,
  Man, Cybern. 17~(4) (1987) 689--696.

\bibitem{priestley72nonparametric}
M.~B. Priestley, M.-T. Chao, Non-parametric function fitting, J. Roy. Statistal
  Soc.: Ser. B (Methodol.) 34~(3) (1972) 385--392.

\bibitem{wilson97advances}
D.~R. Wilson, Advances in instance-based learning algorithms, Ph.D. thesis,
  Brigham Young University (1997).

\bibitem{wilson00integrated}
D.~R. Wilson, T.~R. Martinez, An integrated instance-based learning algorithm,
  Comput. Intell. 16~(1) (2000) 1--28.

\bibitem{hechenbichler04weighted}
K.~Hechenbichler, K.~Schliep, Weighted $k$-nearest-neighbor techniques and
  ordinal classification, Sonderforschungsbereich 386, paper 399,
  Ludwig-Maximilians-Universität München, Institut für Statistik (2004).

\bibitem{derrac14fuzzy}
J.~Derrac, S.~Garc{\'\i}a, F.~Herrera, Fuzzy nearest neighbor algorithms:
  Taxonomy, experimental analysis and prospects, Inf. Sci. 260 (2014) 98--119.

\bibitem{dubois90rough}
D.~Dubois, H.~Prade, Rough fuzzy sets and fuzzy rough sets, Int. J. General
  Syst. 17~(2-3) (1990) 191--209.

\bibitem{pawlak81rough}
Z.~Pawlak, Rough sets, Report 431, ICS PAS (1981).

\bibitem{cornelis10ordered}
C.~Cornelis, N.~Verbiest, R.~Jensen, Ordered weighted average based fuzzy rough
  sets, in: Proc. 5th Int. Conf. Rough Set Knowl. Technol., 2010, pp. 78--85.

\bibitem{verbiest12seleccion}
N.~Verbiest, C.~Cornelis, F.~Herrera, Selecci{\'o}n de prototipos basada en
  conjuntos rugosos difusos, in: 16. Congreso Espa{\~n}ol sobre
  Tecnolog{\'\i}as y L{\'o}gica Fuzzy, 2012, pp. 638--643.

\bibitem{verbiest14fuzzy}
N.~Verbiest, Fuzzy rough and evolutionary approaches to instance selection,
  Ph.D. thesis, Universiteit Gent (2014).

\bibitem{vluymans19weight}
S.~Vluymans, N.~Mac~Parthal{\'a}in, C.~Cornelis, Y.~Saeys, Weight selection
  strategies for ordered weighted average based fuzzy rough sets, Inf. Sci. 501
  (2019) 155--171.

\bibitem{sarkar07fuzzyrough}
M.~Sarkar, Fuzzy-rough nearest neighbor algorithms in classification, Fuzzy
  Sets and Systems 158~(19) (2007) 2134--2152.

\bibitem{wettschereck94study}
D.~Wettschereck, A study of distance-based machine learning algorithms, Ph.D.
  thesis, Oregon State University (1994).

\bibitem{geler16comparison}
Z.~Geler, V.~Kurbalija, M.~Radovanovi{\'c}, M.~Ivanovi{\'c}, Comparison of
  different weighting schemes for the {$k$NN} classifier on time-series data,
  Knowl. Inf. Syst. 48 (2016) 331--378.

\bibitem{geler20weighted}
Z.~Geler, V.~Kurbalija, M.~Ivanovi{\'c}, M.~Radovanovi{\'c}, Weighted {$k$NN}
  and constrained elastic distances for time-series classification, Expert
  Syst. Appl. 162, art. no. 113829 (2020).

\bibitem{higashi82measures}
M.~Higashi, G.~J. Klir, On measures of fuzziness and fuzzy complements, Int. J.
  General Syst. 8~(3) (1982) 169--180.

\bibitem{lukasiewicz23interpretacja}
J.~{\L}ukasiewicz, Interpretacja liczbowa teorii zda{\'n}, Ruch Filozoficzny
  7~(6) (1923) 92--93.

\bibitem{zadeh65fuzzy}
L.~A. Zadeh, Fuzzy sets, Information and control 8~(3) (1965) 338--353.

\bibitem{yager80general}
R.~R. Yager, On a general class of fuzzy connectives, Fuzzy Sets Syst. 4~(3)
  (1980) 235--242.

\bibitem{benjamini95controlling}
Y.~Benjamini, Y.~Hochberg, Controlling the false discovery rate: a practical
  and powerful approach to multiple testing, Journal of the Royal Statistical
  Society: Series B (Methodological) 57~(1) (1995) 289--300.

\bibitem{zhang15crossvalidation}
Y.~Zhang, Y.~Yang, Cross-validation for selecting a model selection procedure,
  Journal of Econometrics 187~(1) (2015) 95--112.

\bibitem{bengio04no}
Y.~Bengio, Y.~Grandvalet, No unbiased estimator of the variance of k-fold
  cross-validation, Journal of Machine Learning Research 5 (2004) 1089--1105.

\bibitem{bousquet02stability}
O.~Bousquet, A.~Elisseeff, Stability and generalization, The Journal of Machine
  Learning Research 2 (2002) 499--526.

\bibitem{malherbe17global}
C.~Malherbe, N.~Vayatis, Global optimization of {L}ipschitz functions, in: ICML
  2017: Proceedings of the 34th International Conference on Machine Learning,
  Vol.~70 of Proceedings of Machine Learning Research, 2017, pp. 2314--2323.

\bibitem{powell09bobyqa}
M.~J.~D. Powell, The {BOBYQA} algorithm for bound constrained optimization
  without derivatives, Tech. Rep. NA2009/06, University of Cambridge,
  Department of Applied Mathematics and Theoretical Physics (2009).

\bibitem{king17global}
D.~E. King, A global optimization algorithm worth using,
  \url{http://blog.dlib.net/2017/12/a-global-optimization-algorithm-worth.html},
  last accessed 6 Jan 2021 (Dec. 2017).

\bibitem{king09dlibml}
D.~E. King, Dlib-ml: A machine learning toolkit, Journal of Machine Learning
  Research 10~(60) (2009) 1755--1758.

\bibitem{lenz22optimised}
O.~U. Lenz, D.~Peralta, C.~Cornelis, Optimised one-class classification
  performance, Machine Learning 111~(8) (2022) 2863--2883.

\bibitem{sugeno73constructing}
M.~Sugeno, Constructing fuzzy measure and grading similarity of patterns by
  fuzzy integral, Trans. Soc. Instrum. Control Engineers 9~(3) (1973) 361--368.

\end{thebibliography}

\end{document}